\documentclass[11pt]{article}
\usepackage[a4paper,margin=25mm]{geometry}
\usepackage{amsmath,amssymb,bm,booktabs,graphicx,xcolor,longtable,array,tabularx}
\newcolumntype{L}[1]{>{\raggedright\arraybackslash\hspace{0pt}}p{#1}}
\usepackage[hyphens]{url}
\usepackage[numbers,sort&compress]{natbib}
\usepackage[section]{placeins}
\usepackage[hidelinks]{hyperref}
\usepackage{authblk}

\graphicspath{{figures/}}

\title{Transferring local rib responses to new component geometries in sheet forming with a graph-based Transformer}
\author[1]{Yingxue Zhao}
\author[1]{Haoran Li}
\author[1]{Haosu Zhou}
\author[2]{Tobias Pfaff}
\author[1]{Nan Li\thanks{Corresponding author. E-mail address: n.li09@imperial.ac.uk (N. Li)}}
\affil[1]{Dyson School of Design Engineering, Imperial College London, London, UK}
\affil[2]{Prometheus, UK}
\date{}

\begin{document}
\maketitle

\begin{abstract}
Sheet metal components often consist of a global geometry, such as the corner radii and forming depth of a corner component, and local features added to it, such as ribs, beads and pads.
During forming, the local features and the global geometry together affect the material flow.
A surrogate that learns all their combinations may require a large amount of training data.
However, the same type of local feature is often used on many global geometries, and its effect may be similar but not identical between them.
Therefore, this study partly decouples the local features from the global geometry.
The effect of the local features is first learned on one global geometry, which requires fewer simulations than learning all combinations.
It is then transferred to other global geometries through transfer learning, which can require fewer target simulations than training from scratch.
A baseline-conditioned residual surrogate, R-BiLoGo, is proposed for ribs on the side walls of hot-stamped corner components.
Its inputs are the baseline response, which is the Finite Element (FE) result of the same component without ribs, and the geometric perturbation, which describes the change in the tool surface caused by the ribs.
Its outputs are the changes in displacement and thinning from the baseline response.
R-BiLoGo gave lower errors than a hierarchical graph-based model and a Transolver-based model on the source geometry, and on two of three selected target geometries after transfer.
Across the 22 target geometries of the systematic study, transfer learning with 20 target simulations gave errors similar to or lower than those of training from scratch with 80 target simulations for about half of the target geometries.
With the same 80 target simulations, transfer learning reduced the errors for 21 of the 22 target geometries.
The effect of the global geometry on transfer was studied with four parameter-free geometric descriptors and controlled variations of standard, chamfered, elliptical and ogee corners.
The benefit of transfer decreased with increasing surface distance and high-curvature change between the source and target geometries, and for targets shallower than the source.
These trends were also observed in other corner types with a different magnitude, and the forming-depth change showed the most consistent effect.
\end{abstract}

\noindent\textbf{Keywords:} Sheet forming; Transfer learning; Graph neural network; Neural operator; Surrogate model; Local geometric features

\section{Introduction}
\label{sec:res_intro}

Sheet metal components often consist of a global geometry and local features added to it, and the two together affect the material flow during forming.
Here, the global geometry refers to the main shape of a component, such as the corner radii and the forming depth of a corner component.
The local features refer to small features added to the global geometry, such as ribs, beads and pads \cite{Attar2023Implicit}.
For example, beads and ribs are added to sheet panels to increase their stiffness \cite{Yang1996BeadPattern}.
During forming, the corner radii and the forming depth affect how the material flows into the die \cite{Horton2020CornerRadii,Attar2021DesignGuidelines}.
A rib changes the material flow around it, and it can change the deformation and thinning of the surrounding region.
Therefore, the same rib may cause different deformation and thinning on a deep corner component and on a shallow one.

Data-driven surrogates have been developed to predict the forming results faster than Finite Element (FE) simulations.
FE simulations are commonly used to predict the deformation and thinning of formed components, but running them for many design candidates can be time-consuming.
Surrogates trained on FE results can predict the deformation and thinning rapidly once trained \cite{Attar2021Rapid,Attar2023Platform,Xu2022SuperMeshing,Zhou2025BlankShape,Mitoubsi2026FormingGNN,Baum2025Forming3DSurrogate,Wang2024TubeGNN,Brzobohaty2026IncrementalForming}.
Among them, Graph Neural Networks (GNNs) and Transformer-based models can predict the displacement at each node and the thinning at each element of the FE mesh \cite{Pfaff2021MeshGraphNets,Wu2024Transolver,Zhao2024GNNReview,Zhou2025MultiLevel,Zhao2025BatteryBox,Zhao2026RUGNN,Zhao2026CAttBiGNN,Zhao2026BiLoGo}.
Similar surrogates have also been developed to predict the crashworthiness of formed components \cite{Li2024CrashCNN,Li2026CrashGNN,Li2026MaskMorph}.
In these models, the input is the geometry of the component, and the output is the complete displacement and thinning fields.

However, a surrogate that learns all combinations of global geometry and local features may require a large amount of training data.
In most studies, either the global geometry is changed by a few parameters, such as the corner radii and the forming depth \cite{Attar2021Rapid,Zhao2026RUGNN,Zhao2026CAttBiGNN}, or the local features are changed on one fixed global geometry \cite{Zhao2026BiLoGo}.
Even when only the global geometry and the material were changed, 3000 FE simulations were used for one material system in \cite{Luo2026StampFormer}.
If the size and position of the local features are also changed, the training data would need to cover their combinations with the global geometry.
In addition, when a new global geometry is introduced, the surrogate may need to be retrained.

In practice, the same type of local feature is often used on many global geometries, and its effect may be similar but not identical between them.
Since the effects are similar, the knowledge learned on one global geometry may be reused on another.
Since the effects are not identical, this knowledge needs to be adapted to the new global geometry.
Therefore, the local features can be partly decoupled from the global geometry.
The effect of the local features is first learned on one global geometry, which requires fewer simulations than learning all combinations.
It is then transferred to other global geometries with a small number of simulations.

Transfer learning can reduce the data required for a new task, but its benefit is not guaranteed.
In transfer learning, a model trained on a source task is reused for a target task \cite{Pan2010TransferSurvey,Zhuang2021TransferSurvey}.
In structural mechanics, graph-based surrogates have been transferred between truss structures with different geometries \cite{Whalen2022ReusableTrusses}.
In forming, pretrained models have been transferred to new incremental-forming processes \cite{Duan2024DSIFTransfer}, from analytical forming-force data to deep-drawing simulations \cite{Guo2025DeepDrawingTransfer}, and between tube-bending processes \cite{Wang2025CrossFormingTransfer}.
These forming studies transferred the model to a new process or a new type of data, rather than to a new global geometry with local features.
If the source and target tasks are too different, the transferred model can have a higher error than a model trained only on the target data, which is referred to as negative transfer \cite{Zhang2023NegativeTransfer,Li2022NegativeTransferGP}.
Therefore, it is important to choose what the model learns on the source geometry, so that this knowledge can be reused on a new global geometry.

In addition, the benefit of transfer can be different for different target geometries, and it is useful to know which geometric differences reduce it.
Such knowledge could help to decide whether to transfer the model before new simulations are run.
In machine learning, methods have been proposed to estimate whether a pretrained model is suitable for a new task \cite{Xue2024TransferabilityReview}.
In structural health monitoring, the benefit of transfer between structures has been related to measures of their similarity \cite{Bull2021PBSHMI,Gardner2022BayesianTransfer,Brennan2025StructuralSimilarity,Giglioni2026SimilarityRequirements}.
For surrogates of sheet forming, how the benefit of transfer changes with the global geometry has been less studied.

To address these problems, this study proposes a baseline-conditioned residual surrogate, named R-BiLoGo, for ribs on the side walls of hot-stamped corner components.
For each global geometry, one FE simulation is run without ribs, and its result is referred to as the baseline response.
The inputs of the surrogate are the baseline response and the geometric perturbation, which describes the change in the tool surface caused by the ribs.
The outputs are the changes in displacement and thinning from the baseline response, which are referred to as the residual fields.
The baseline response is used as an input because the effect of the ribs depends on the global geometry.
A similar approach is used in multi-fidelity modelling, in which a model learns the correction to a low-fidelity prediction \cite{Kennedy2001Calibration,Chakraborty2021MultifidelityTransfer,Conti2023MultifidelityLSTM}.
R-BiLoGo extends BiLoGo \cite{Zhao2026BiLoGo} with two separate streams for the baseline response and the geometric perturbation.
The model is first trained on one corner geometry with 193 rib configurations, and then transferred to new corner geometries with a small number of simulations.
The effect of the global geometry on transfer is then studied systematically on target geometries with standard, chamfered, elliptical and ogee corners.

The contributions of this study are as follows.
First, a baseline-conditioned residual formulation is defined for ribs on hot-stamped corner components, in which the residual displacement and thinning fields are predicted in one forward pass.
The ablation study shows that the residual targets mainly reduce the node error, while the baseline response as input and the baseline fusion mainly reduce the thinning error.
Second, R-BiLoGo gave lower errors than a hierarchical graph-based model and a Transolver-based model on the source geometry, and on two of three selected target geometries after transfer.
Across the 22 target geometries of the systematic study, transfer learning with 20 target simulations gave errors similar to or lower than those of training from scratch with 80 simulations for about half of the target geometries, and with the same 80 simulations, transfer learning reduced the errors for 21 of the 22 target geometries.
Third, the effect of the global geometry on transfer is studied with four parameter-free geometric descriptors.
Three descriptors were selected from 28 candidates by their behaviour on 430 CAD geometries, and the fourth was introduced from the transfer results.
Target geometries were generated by changing one descriptor while limiting the changes in the others.
The forming-depth change showed the most consistent effect across the subclasses, while the maximum surface distance and the change in the high-curvature regions showed the same direction with a magnitude that depends on the subclass.
Therefore, these descriptors can indicate target geometries with a lower benefit of transfer, and the relationships may be calibrated for a new subclass with a small number of target geometries.

The remainder of this study is organised as follows.
Section~\ref{sec:res_formulation} formulates the problem.
Section~\ref{sec:res_dataset} describes the forming simulations, the geometries and the datasets.
Section~\ref{sec:res_model} presents the surrogate.
Section~\ref{sec:res_source_model} compares R-BiLoGo with two baseline models on the source dataset and after transfer learning with 20 target simulations, and presents an ablation study.
Section~\ref{sec:res_transfer} examines the effect of the number of target simulations on transfer learning.
Section~\ref{sec:res_systematic} presents the systematic study of the effect of the global geometry on transfer, and Section~\ref{sec:res_summary} gives the conclusions.


\section{Problem formulation}
\label{sec:res_formulation}

This section defines the residual fields predicted by the surrogate, the inputs required to predict them, and the transfer learning setting.

The residual fields describe the change in the forming result when ribs are added to a component, while its global geometry is kept unchanged.
They are obtained by comparing two FE simulations of the same global geometry.
The first simulation uses the tools with the ribs, and it is referred to as the rib simulation.
The second simulation uses the same tools without the ribs, and it is referred to as the baseline simulation.
Both simulations use the same blank mesh, material, process conditions and friction coefficient.
Therefore, each node and each element of the blank has a counterpart in the other simulation, and the two results can be compared node by node and element by element.

The residual displacement of node $i$ at the final state is defined as
\begin{equation}
\Delta\bm{u}_i = \bm{x}_i^{\mathrm{rib}} - \bm{x}_i^{\mathrm{base}},
\label{eq:res_disp}
\end{equation}
where $\bm{x}_i^{\mathrm{rib}}$ and $\bm{x}_i^{\mathrm{base}}$ denote the final positions of node $i$ in the rib and baseline simulations, respectively.
Since both simulations start from the same initial blank, $\Delta\bm{u}_i$ is also the difference between the nodal displacements $\bm{u}_i^{\mathrm{rib}}$ and $\bm{u}_i^{\mathrm{base}}$ of the two simulations.
The residual thinning of element $j$ is defined as
\begin{equation}
\Delta\theta_j = \theta_j^{\mathrm{rib}} - \theta_j^{\mathrm{base}},
\label{eq:res_thin}
\end{equation}
where $\theta_j^{\mathrm{rib}}$ and $\theta_j^{\mathrm{base}}$ denote the final thinning of element $j$ in the rib and baseline simulations, respectively.
The thinning is defined as $\theta = 1 - t/t_0 = 1 - \exp(\varepsilon_{33})$, where $t$ and $t_0$ denote the current and initial thickness, and $\varepsilon_{33}$ denotes the logarithmic strain in the thickness direction.
A positive $\theta$ indicates thinning.
The forming result with the ribs is obtained by adding the predicted residual fields to the baseline response.

\begin{figure}[!htbp]
\centering
\includegraphics[width=\textwidth]{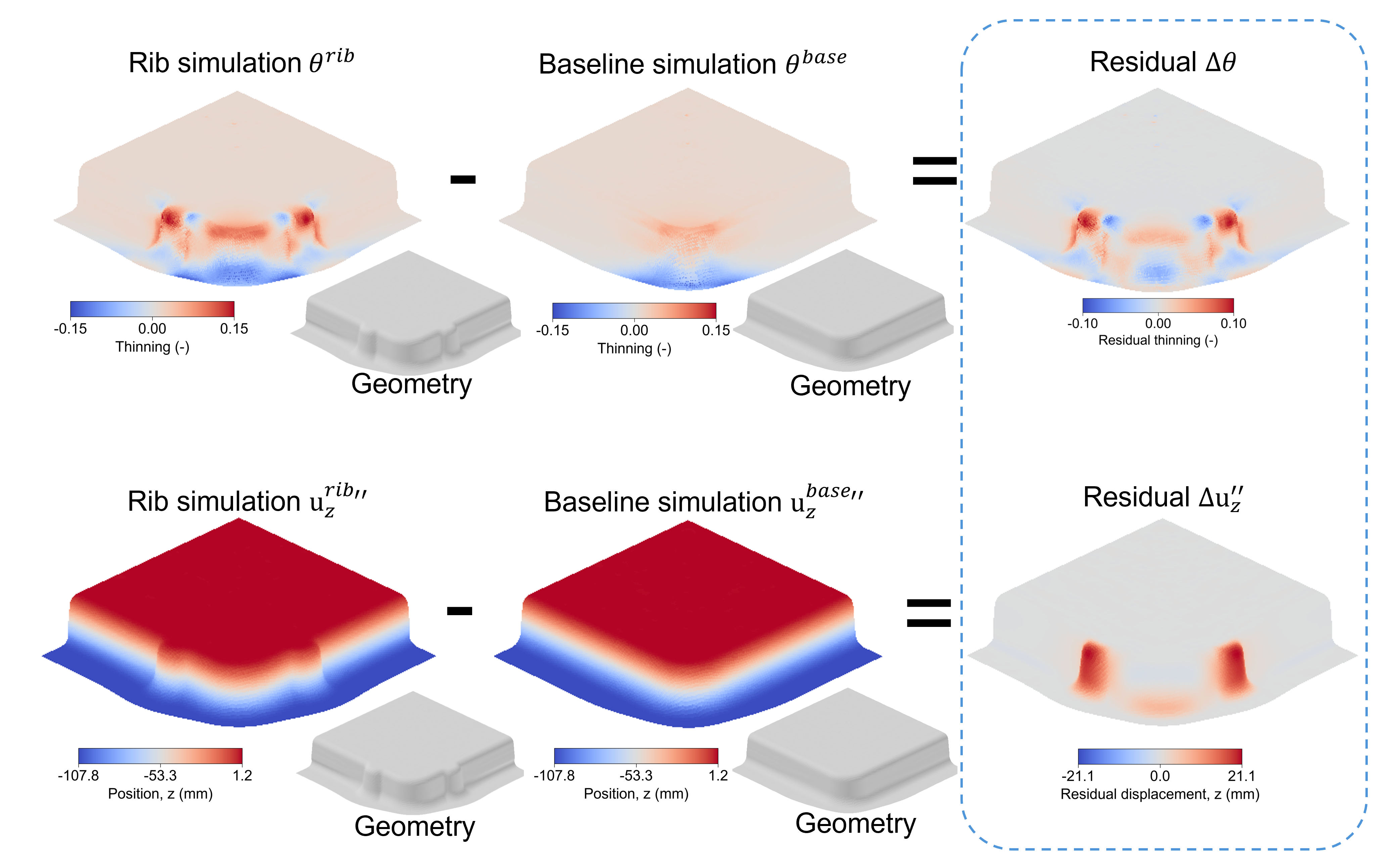}
\caption{Definition of the residual fields for one rib configuration on a fixed global geometry. Top: final thinning of the rib simulation and the baseline simulation, and the residual thinning. Bottom: the same for the displacement, with the $z$ direction shown as an example. The tool geometries of the two simulations are shown in grey. The dashed box marks the residual fields, which are concentrated near the ribs, while the complete fields are dominated by the forming of the global geometry.}
\label{fig:res_residual_definition}
\end{figure}

An example of the residual fields is shown in Fig.~\ref{fig:res_residual_definition}.
In this example, the largest residuals occur near the ribs, while smaller changes also appear in the surrounding region, such as the bottom of the corner.
The residual displacement in the $z$ direction reaches about 21~mm near the ribs, while the complete displacement in the same direction reaches about 108~mm.

\begin{figure}[!htbp]
\centering
\includegraphics[width=\textwidth]{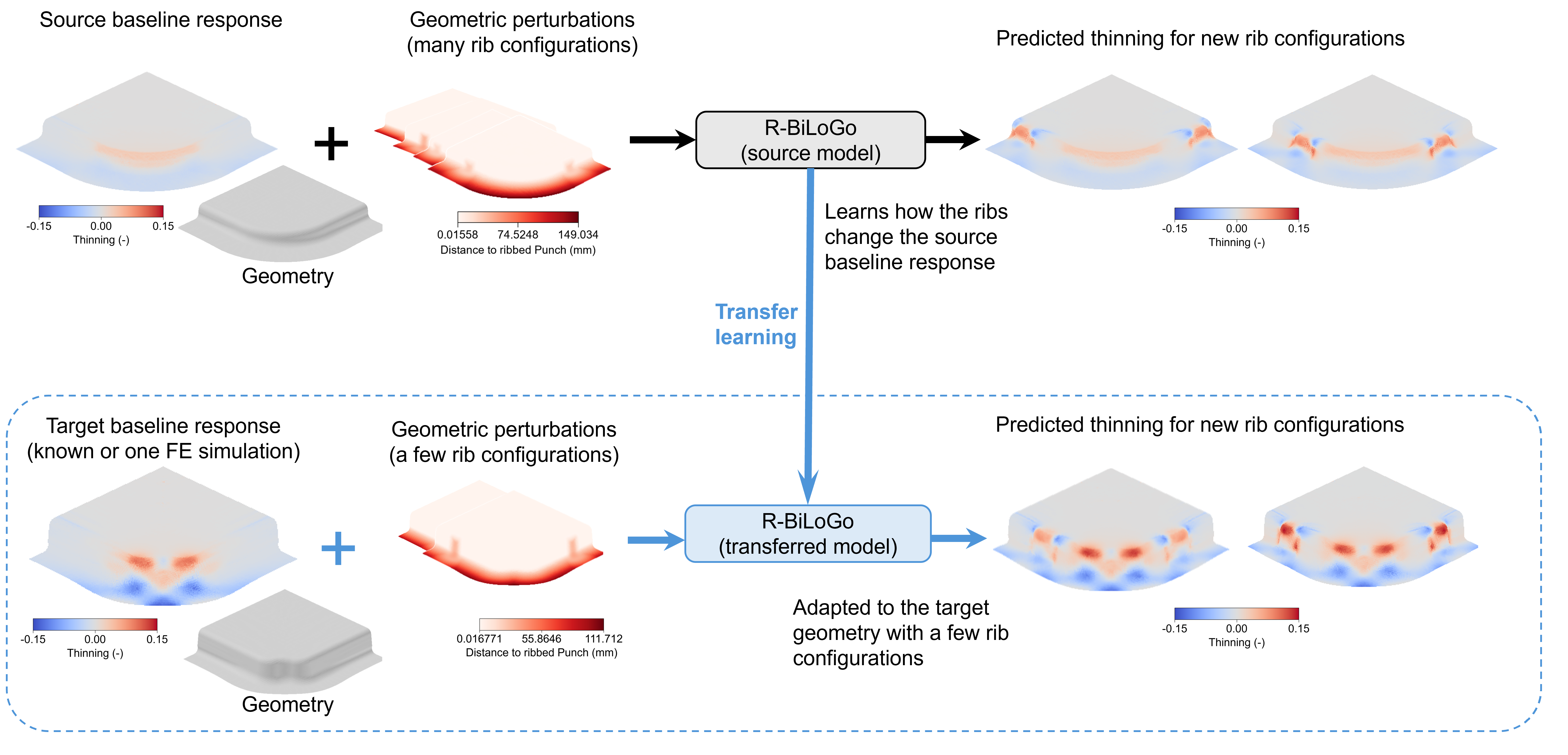}
\caption{Transfer workflow. The source model is trained with the baseline response of the source geometry and the geometric perturbations of many rib configurations. It is then transferred to a target geometry with a different corner shape, using its baseline response and a few rib configurations. Each target geometry is transferred separately. Although the baseline responses of the two geometries are different, the changes near the ribs show a similar pattern.}
\label{fig:res_workflow}
\end{figure}

The residual fields cannot be predicted from the rib geometry alone.
The ribs change the contact between the blank and the tools, and this change may depend on how the blank deforms in the baseline simulation.
Therefore, the same rib may cause different residuals on different global geometries.
This can be seen in Fig.~\ref{fig:res_workflow}, in which the changes near the ribs show a similar pattern on two global geometries with different corner shapes, but with different magnitudes and surrounding regions.
This dependence is also examined quantitatively by the zero-shot prediction in Section~\ref{sec:res_selected}.
For this reason, the surrogate receives two inputs.
The first input is the baseline response, which is the result of the baseline simulation.
The second input is the geometric perturbation, which is described by the distances and surface normals from the baseline final state to the tools with ribs (Section~\ref{sec:res_model}).
With both inputs, the surrogate receives the deformed state of the global geometry together with the change in the tools, and this state is replaced by that of the new global geometry when the model is transferred.
The surrogate $f$ is trained to approximate the mapping
\begin{equation}
\left(\{\Delta\bm{u}_i\},\ \{\Delta\theta_j\}\right) = f\left(\mathcal{B},\ \mathcal{P}\right),
\label{eq:res_mapping}
\end{equation}
where $\{\Delta\bm{u}_i\}$ and $\{\Delta\theta_j\}$ denote the residual displacements of all nodes and the residual thinning of all elements, and $\mathcal{B}$ and $\mathcal{P}$ denote the baseline response and the geometric perturbation, respectively.

The formulation also determines what is reused in transfer learning and how the surrogate is used for a new global geometry.
The global geometry used to train the surrogate is referred to as the source geometry, and a new global geometry is referred to as a target geometry.
On the source geometry, $f$ is trained with $N_{\mathrm{src}}$ rib simulations, and it learns how the ribs change a given baseline response.
For a target geometry, the baseline response is often available.
Otherwise, it can be obtained from one FE simulation.
The surrogate is then initialised from the source model and trained with a small number $K$ of rib simulations of the target geometry, in which ribs of different positions and sizes are used.
After transfer learning, the surrogate can predict the displacement and thinning caused by ribs of different positions and sizes on the target geometry.
This workflow is shown in Fig.~\ref{fig:res_workflow}.

\section{Datasets}
\label{sec:res_dataset}

In this study, the global geometry refers to the main shape of the formed part, which is defined by the die radius, the punch radius, the forming depth and the shape of the corner in the plan view.
The local features are a pair of reinforcing ribs added to the side walls of this global geometry.
In all datasets, the global geometry and the ribs are changed, while the material and the process conditions are kept the same.
This section describes the simulation setup, the global geometries and the ribs, and the source and target datasets.

\subsection{Simulation setup}
\label{sec:res_simulation}

\begin{figure}[!htbp]
\centering
\includegraphics[width=\textwidth]{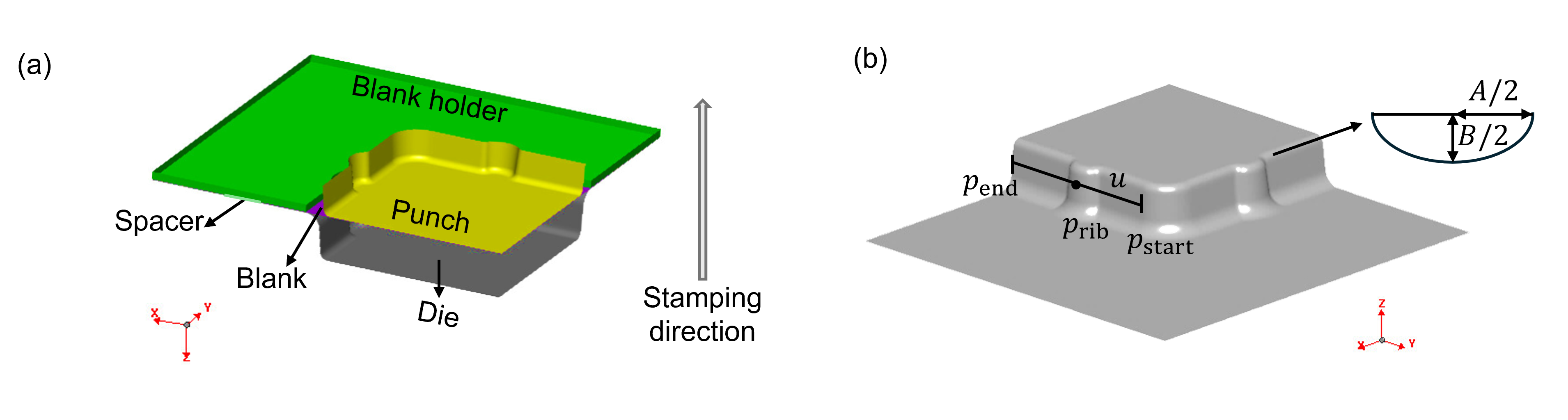}
\caption{(a) Example of the simulation setup for one global geometry, with the blank and the tools of the quarter model. (b) Rib parameters on the formed part. The FE setup and the rib parameterisation are the same as in \cite{Zhao2026BiLoGo}.}
\label{fig:res_fe_setup}
\end{figure}

All datasets are obtained from FE simulations of the hot stamping of a deep-drawn corner, which is a typical feature of battery box components \cite{Zhao2025BatteryBox,Zhao2026CAttBiGNN}.
The blank is made of AA6082 with an initial thickness of 2~mm, and it is described by an elastic--viscoplastic material model for hot-stamping conditions \cite{Mohamed2012Investigation,Mohamed2015CDM,Zhao2026RUGNN}.
The blank is heated to 500~$^{\circ}$C, and the tools are kept at 25~$^{\circ}$C.
The stamping speed is 500~mm/s, the friction coefficient is 0.1 and the blank-holder force is 200~kN.
The simulations were performed in PAM-STAMP with a quarter model, as shown in Fig.~\ref{fig:res_fe_setup}(a).
The blank is discretised with 16,049 nodes and 31,595 three-node triangular shell elements, and the same blank mesh is used in all simulations.

\subsection{Global geometries and ribs}
\label{sec:res_geometries}

\begin{figure}[!htbp]
\centering
\includegraphics[width=\textwidth]{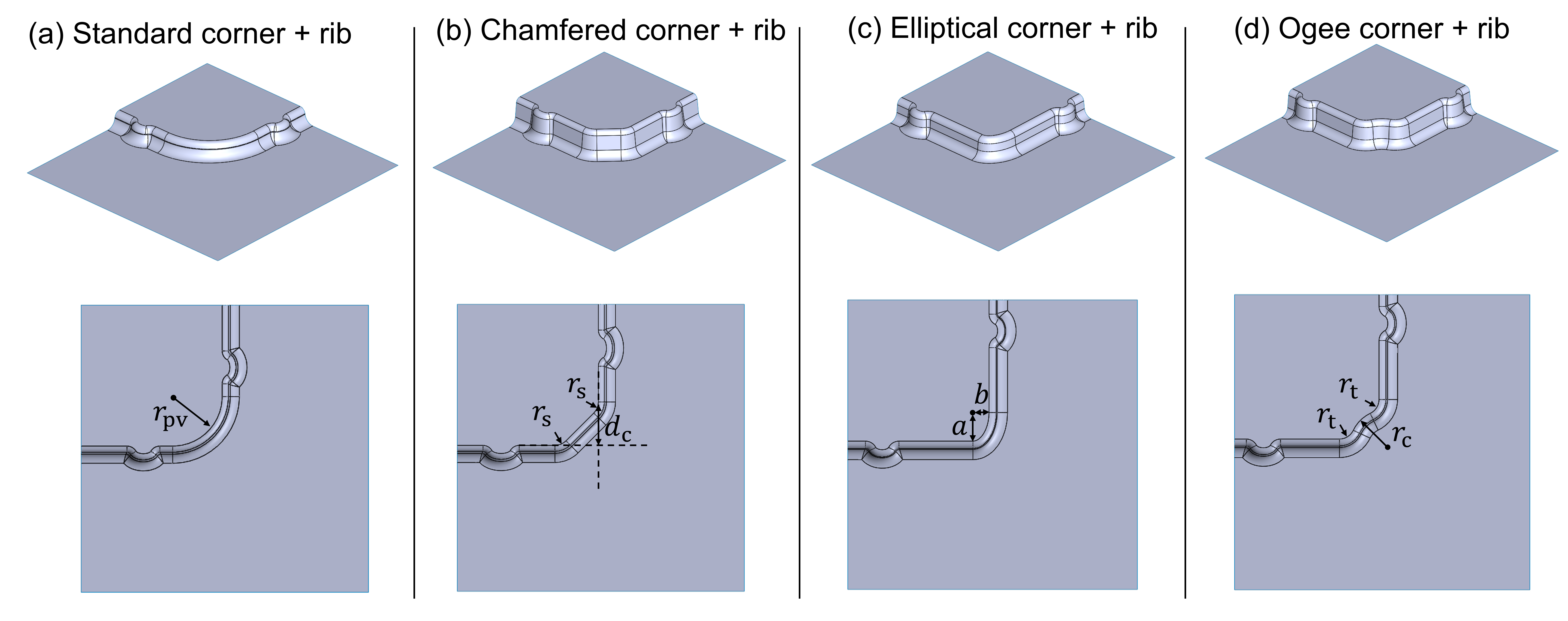}
\caption{The four corner-geometry subclasses with ribs: (a) Standard, (b) Chamfered, (c) Elliptical and (d) Ogee. The top row shows the forming surfaces, and the bottom row shows the plan view of the corner with the subclass-specific parameters.}
\label{fig:geometry_subclasses}
\end{figure}

The global geometries are corner geometries from four subclasses, which differ in the shape of the corner in the plan view.
The four subclasses are Standard, Chamfered, Elliptical and Ogee, as shown in Fig.~\ref{fig:geometry_subclasses}.
All four subclasses share the die radius $r_{\mathrm d}$, the punch radius $r_{\mathrm p}$ and the forming depth $H$.
The Standard subclass has a circular corner with the plan-view radius $r_{\mathrm{pv}}$.
The Chamfered subclass replaces the circular corner with a straight chamfer. Here, $d_{\mathrm c}$ is the distance from the intersection of the extended side walls to the end of the corner, measured along the side wall. The two ends of the chamfer are connected to the straight side walls by fillets of radius $r_{\mathrm s}$.
The Elliptical subclass uses an elliptical corner. Here, $a$ and $b$ are the semi-axes of the elliptical corner in the vertical and horizontal directions of Fig.~\ref{fig:geometry_subclasses}(c), respectively.
The Ogee subclass uses a corner that consists of a central arc of radius $r_{\mathrm c}$, which is connected to the two straight side walls by two arcs of radius $r_{\mathrm t}$ with the opposite curvature.
The parameter ranges of each subclass are listed in Table~\ref{tab:geometry_subclass_parameters}.

\begin{table}[!htbp]
\centering
\caption{Parameter ranges of the four corner-geometry subclasses. All lengths are in mm. For the Elliptical subclass, $b$ and the ratio $a/b$ were sampled, and $a$ was calculated from them.}
\label{tab:geometry_subclass_parameters}
\small
\begin{tabular}{@{}lcccl@{}}
\toprule
Subclass & $r_{\mathrm d}$ & $r_{\mathrm p}$ & $H$ & Subclass-specific parameters \\
\midrule
Standard & 25--40 & 20--27.3 & 60--120 & $r_{\mathrm{pv}}$: 70--250 \\
Chamfered & 25--40 & 20--27.3 & 60--120 & $r_{\mathrm s}$: 70--100; $d_{\mathrm c}$: 70--140 \\
Elliptical & 25--40 & 20--27.3 & 60--120 & $b$: 65--100; $a/b$: 1.20--1.75 \\
Ogee & 25--40 & 20--27.3 & 60--120 & $r_{\mathrm c}$: 80--120; $r_{\mathrm t}$: 50--80 \\
\bottomrule
\end{tabular}
\end{table}

On each global geometry, the ribs are placed on the two side walls next to the corner.
In the top view, each rib has the shape of half an ellipse, with its straight edge on the side wall.
The rib geometry is defined by three parameters, as listed in Table~\ref{tab:res_rib_parameters}.
The position parameter $u$ is the distance from $p_{\mathrm{start}}$ to the rib centre $p_{\mathrm{rib}}$ along the straight top edge of the side wall, given as a percentage of the edge length from $p_{\mathrm{start}}$ to $p_{\mathrm{end}}$, as shown in Fig.~\ref{fig:res_fe_setup}(b).
The parameters $A/2$ and $B/2$ are the semi-axes of the rib along and normal to the side wall.
In all datasets, the rib parameters were sampled by Latin Hypercube Sampling (LHS) \cite{McKay1979LHS} in the ranges of Table~\ref{tab:res_rib_parameters}.
Since $u$ is given as a percentage of the edge length, the same ranges can be used for all global geometries.
For each global geometry, one FE simulation without ribs was performed to provide the baseline response.

\begin{table}[!htbp]
\centering
\caption{Rib parameters of the source dataset \cite{Zhao2026BiLoGo}.}
\label{tab:res_rib_parameters}
\small
\begin{tabular}{@{}llp{7cm}@{}}
\toprule
Parameter & Range & Description \\
\midrule
$u$ & 25--75\% & Position of the rib centre along the top edge of the side wall \\
$A/2$ & 35--50~mm & Half-length of the rib along the side wall \\
$B/2$ & 22.5--30~mm & Outward extent of the rib from the side wall \\
\bottomrule
\end{tabular}
\end{table}

\subsection{Source and target datasets}
\label{sec:res_datasets}

Four groups of datasets were used: the source dataset, three selected target datasets, the target datasets within the Standard subclass, and the target datasets of the other subclasses.

\paragraph{Source dataset}
The source dataset is the Corner+Rib dataset of \cite{Zhao2026BiLoGo}.
Its global geometry is a Standard corner with $r_{\mathrm d} = 32.5$~mm, $r_{\mathrm p} = 23.65$~mm, $r_{\mathrm{pv}} = 160$~mm and $H = 90$~mm.
These values are at the centre of the parameter ranges of the Standard subclass in Table~\ref{tab:geometry_subclass_parameters}.
The training set contains 193 simulations, and the validation and test sets contain 50 simulations each.
The validation and test configurations were sampled separately by LHS, and they are not part of the training set.

\paragraph{Rib configurations of the target datasets}
For the target geometries, the rib configurations were taken from a separate pool of 294 configurations sampled by LHS.
A larger pool than the source dataset was used so that nested subsets of up to 160 configurations could be selected from it.
Since the source and target configurations were sampled independently, the target simulations contain rib configurations that were not seen during source training.
The validation and test configurations of the target datasets were also sampled separately by LHS, and they are not part of this pool.
To study how many target simulations are required for an accurate prediction after transfer learning, nested subsets of this pool were selected,
\begin{equation}
S_{5} \subset S_{10} \subset S_{20} \subset S_{40} \subset S_{80} \subset S_{160},
\label{eq:res_subsets}
\end{equation}
where $S_K$ denotes the subset with $K$ rib configurations, and $K$ is referred to as the number of target simulations.
Each smaller subset is contained in the larger ones, and each subset covers the rib parameter space evenly.
The subsets were selected only from the rib parameters with a deterministic method based on a Halton sequence \cite{Halton1960Sequence}, as described in Appendix~\ref{app:res_subsets}.

\begin{figure}[!htbp]
\centering
\includegraphics[width=\textwidth]{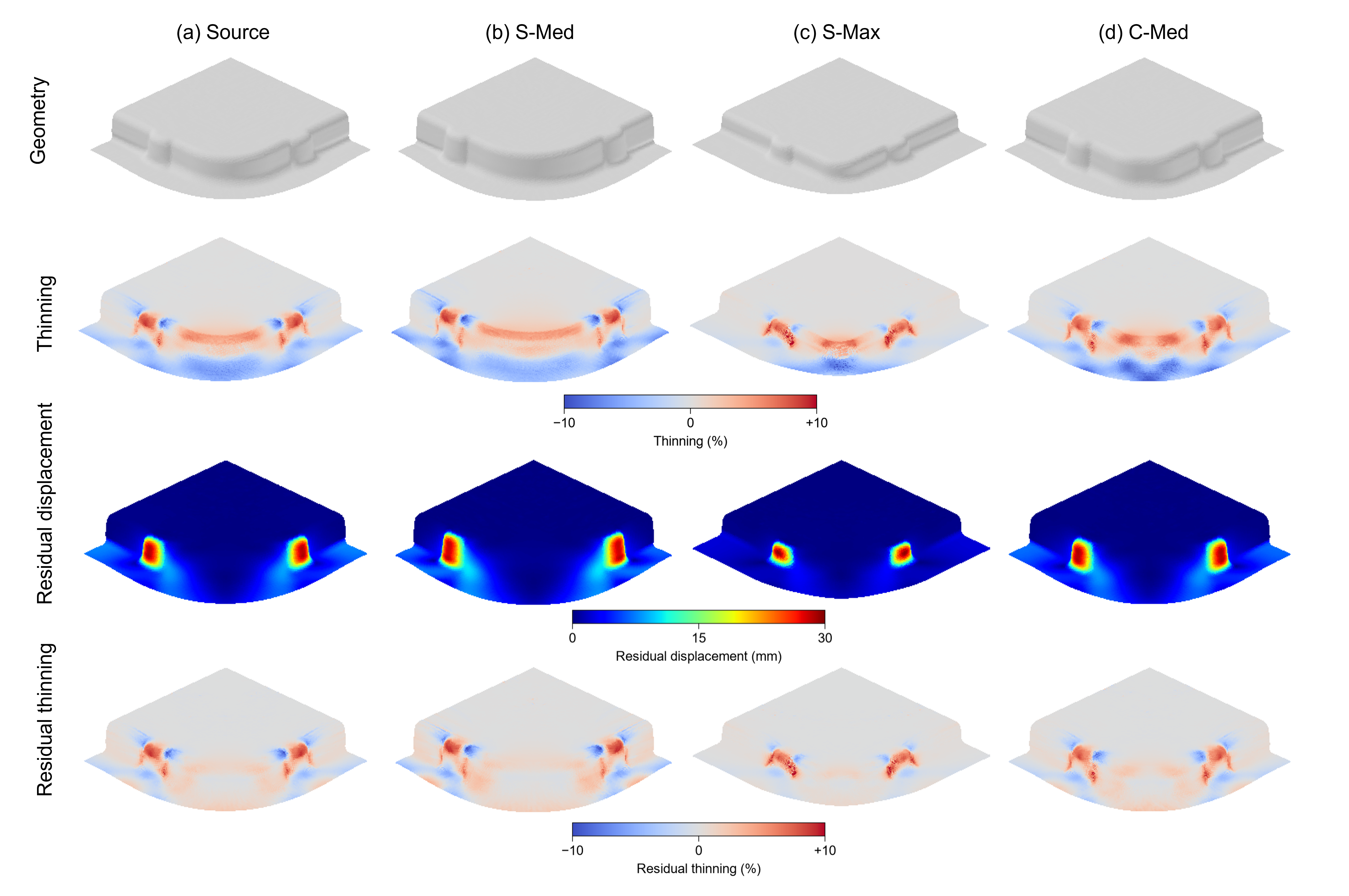}
\caption{Source geometry and the three selected target geometries: (a) source, (b) S-Med, (c) S-Max and (d) C-Med. From top to bottom: global geometry with ribs, FE thinning, magnitude of the residual displacement and residual thinning of an example simulation. The same rib configuration is used on all four geometries.}
\label{fig:res_selected_targets}
\end{figure}

\paragraph{Three selected target datasets}
The effect of $K$ was first studied on three target geometries, before the other target geometries were examined with fixed values of $K$.
These three geometries were also used to compare R-BiLoGo with two baseline models.
They were selected by the distance between the source and target geometries without ribs, which was computed with the Chamfer distance \cite{Barrow1977Chamfer}.
The Chamfer distance is commonly used to measure the difference between two surfaces.
S-Med and S-Max are the geometries with the median and the maximum Chamfer distance to the source geometry among 600 Standard geometries sampled by LHS in the ranges of Table~\ref{tab:geometry_subclass_parameters}.
C-Med is the geometry with the median Chamfer distance to the source geometry among 600 Chamfered geometries sampled in the same way.
The three geometries are shown in Fig.~\ref{fig:res_selected_targets}, together with the source geometry and the FE results of an example simulation.
For comparison, the same rib configuration is used on all four geometries in Fig.~\ref{fig:res_selected_targets}, with the same position parameter $u$ and the same rib size.
In the datasets, the position and size of the ribs vary between the simulations, as described in Section~\ref{sec:res_geometries}.

For these three geometries, all subsets in Eq.~\eqref{eq:res_subsets} were used, and the full pool of 294 simulations was used to train a reference model.
The validation and test sets contain 50 simulations each.

\paragraph{Target datasets within the Standard subclass}
Thirteen Standard target geometries were used to relate the transfer benefit to geometric descriptors of the global geometry.
They were selected within the parameter ranges of the Standard subclass, so that one geometric descriptor changes while the changes in the other descriptors are limited.
The descriptors are introduced in Section~\ref{sec:res_descriptors}, and the method used to generate these geometries is described in Appendix~\ref{app:descriptor_control}.

\paragraph{Target datasets of the other subclasses}
Nine further target geometries, including 3 Chamfered, 4 Elliptical and 2 Ogee geometries, were used to check whether the relationships found in the Standard subclass also hold for other corner shapes.
They were also generated by controlled variations of the descriptors, within the parameter ranges of their subclasses.

For the 22 target geometries of the last two groups, only the subsets $S_{20}$ and $S_{80}$ were used.
Therefore, each of these target datasets contains 80 training simulations, together with 25 validation simulations and 50 test simulations.
The same rib configurations, including the validation and test configurations, were used for all 22 target geometries, and they were placed on the global geometry of each target.
As a result, the datasets of different targets differ only in the global geometry.
In total, 25 target geometries were used: the three selected target geometries in Sections~\ref{sec:res_source_model} and \ref{sec:res_transfer}, and the 22 target geometries of the systematic study in Section~\ref{sec:res_systematic}.

\section{Baseline-conditioned bipartite surrogate}
\label{sec:res_model}

This section presents the surrogate, which is referred to as R-BiLoGo, where R denotes the residual targets.
R-BiLoGo extends BiLoGo \cite{Zhao2026BiLoGo}, and predicts the final residual fields in one forward pass, without an autoregressive rollout.
Section~\ref{sec:res_features} describes the graph representation and the features, and Section~\ref{sec:res_architecture} describes the architecture.

\subsection{Graph representation and features}
\label{sec:res_features}

The blank mesh is represented as a node--element bipartite graph $\mathcal{G} = (\mathcal{V}_{\mathrm{N}}, \mathcal{V}_{\mathrm{E}}, \mathcal{E})$ \cite{Zhao2026BiLoGo}.
Here, $\mathcal{V}_{\mathrm{N}}$ denotes the set of mesh nodes, $\mathcal{V}_{\mathrm{E}}$ denotes the set of elements, and $\mathcal{E}$ denotes the set of edges.
Each edge connects an element to one of its three nodes, so that the edges describe the connectivity of the mesh.
The edges carry no features.
The nodes and the elements are used as two types of entities because the displacement is defined at the nodes and the thinning is defined at the elements, so that each output is predicted on its own entity.
Each node and each element has two groups of features, a baseline group and a perturbation group, which represent the two inputs in Eq.~\eqref{eq:res_mapping}, as listed in Table~\ref{tab:res_features}.

The baseline group, which represents the baseline response $\mathcal{B}$, is taken from the reference and final states of the baseline simulation.
For each node, the baseline features contain the reference position, the final position in the baseline simulation and two boundary flags for the nodes on the two symmetry planes.
For each element, they contain the reference position of the element centre and the final thinning in the baseline simulation.
These features describe how the global geometry deforms without ribs.

The perturbation group, which represents the geometric perturbation $\mathcal{P}$, describes the tools with ribs relative to the baseline final state.
For each node of the baseline final state, the closest points on the die and on the punch with ribs are found on the triangulated tool surfaces.
The perturbation features of the node are the distances to these closest points and the unit normals of the tool surfaces at these points.
The distance describes how far the ribbed tool is from the blank, and the normal describes the orientation of the tool surface, which changes near the ribs.
For each element, the perturbation features contain the mean and the minimum of the node distances and the mean of the node normals over the three nodes of the element, for the die and for the punch.
These features describe the tool geometry at the location of each node, so that the tools do not need to be represented by a separate mesh.

All input features are normalised by the mean and standard deviation of the training data of the source geometry.
The outputs are the residual displacement $\Delta\bm{u}_i$ of each node and the residual thinning $\Delta\theta_j$ of each element, as defined in Eqs.~\eqref{eq:res_disp} and \eqref{eq:res_thin}.

\begin{table}[!htbp]
\centering
\caption{Input and output features of R-BiLoGo.}
\label{tab:res_features}
\small
\begin{tabular}{@{}llp{7.2cm}@{}}
\toprule
Group & Feature & Description \\
\midrule
Node, baseline & $\bm{x}_i^{0}$ & Reference node position \\
 & $\bm{x}_i^{\mathrm{base}}$ & Final node position of the baseline simulation \\
 & $b_i^{x}$, $b_i^{y}$ & Boundary flags for nodes on the symmetry planes $x = 0$ and $y = 0$ \\
Element, baseline & $\bm{c}_j^{0}$ & Reference position of the element centre \\
 & $\theta_j^{\mathrm{base}}$ & Final thinning of the baseline simulation \\
\midrule
Node, perturbation & $d_{i}^{\mathrm{D}}$, $\bm{n}_{i}^{\mathrm{D}}$ & Distance to the die with ribs, and unit normal of the die at the closest point \\
 & $d_{i}^{\mathrm{P}}$, $\bm{n}_{i}^{\mathrm{P}}$ & Distance to the punch with ribs, and unit normal of the punch at the closest point \\
Element, perturbation & $\bar{d}_{j}^{\mathrm{D}}$, $d_{j,\min}^{\mathrm{D}}$, $\bar{\bm{n}}_{j}^{\mathrm{D}}$ & Mean and minimum die distance, and mean die normal over the nodes of the element \\
 & $\bar{d}_{j}^{\mathrm{P}}$, $d_{j,\min}^{\mathrm{P}}$, $\bar{\bm{n}}_{j}^{\mathrm{P}}$ & Mean and minimum punch distance, and mean punch normal over the nodes of the element \\
\midrule
Node, output & $\Delta\bm{u}_i$ & Residual displacement (Eq.~\eqref{eq:res_disp}) \\
Element, output & $\Delta\theta_j$ & Residual thinning (Eq.~\eqref{eq:res_thin}) \\
\bottomrule
\end{tabular}
\end{table}

\subsection{Architecture}
\label{sec:res_architecture}

\begin{figure}[!htbp]
\centering
\includegraphics[width=\textwidth]{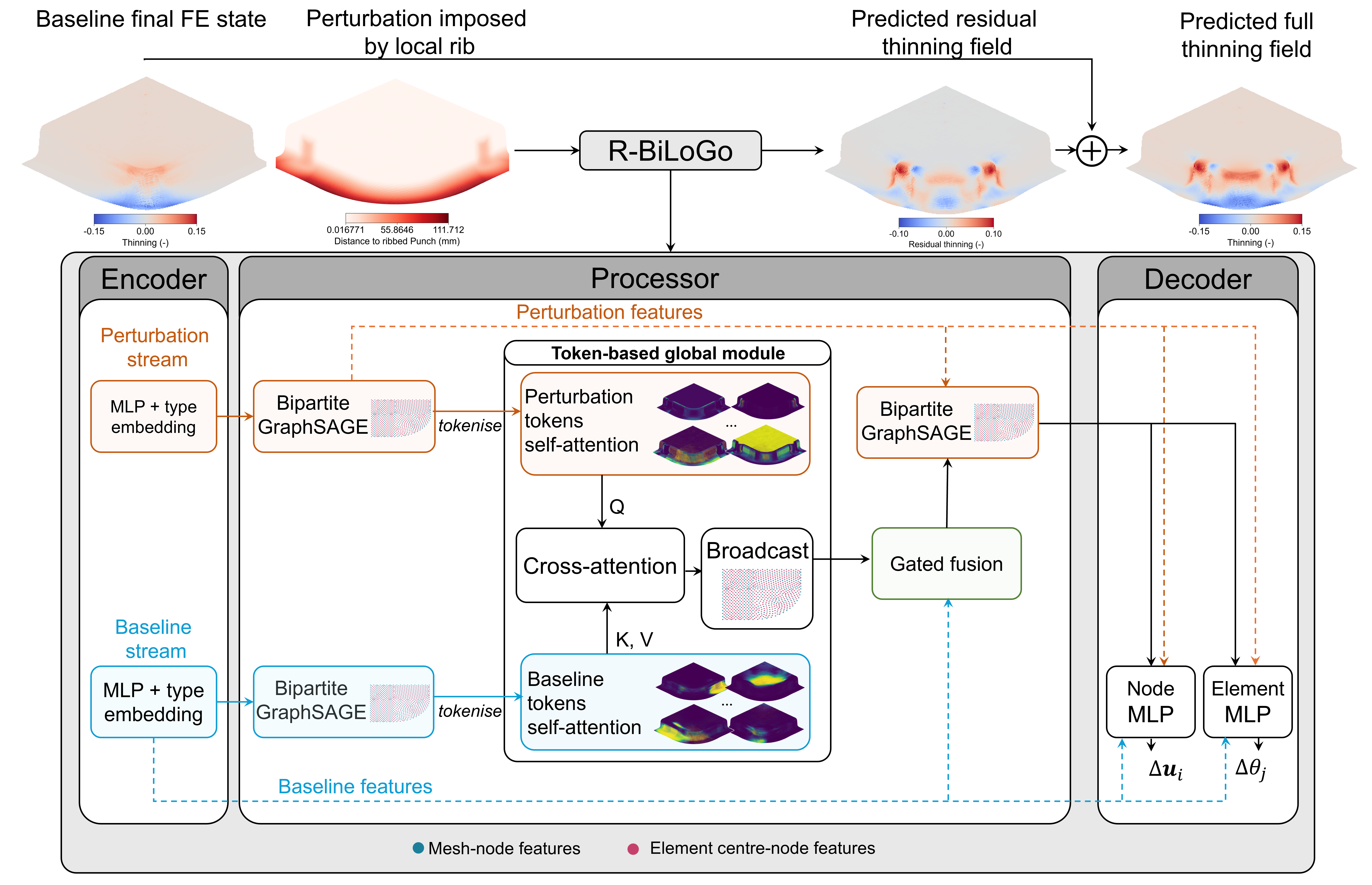}
\caption{Overview and architecture of R-BiLoGo. Top: inputs and outputs of the surrogate, shown for the thinning. Bottom: encoder, processor and decoder. The perturbation stream is shown in orange and the baseline stream in blue. Dashed lines show the features passed directly to later components.}
\label{fig:res_architecture}
\end{figure}

The surrogate follows the encoder--processor--decoder structure of BiLoGo-T \cite{Zhao2026BiLoGo}, as shown in Fig.~\ref{fig:res_architecture}.
The baseline response and the geometric perturbation are processed in two separate streams.
The encoder maps the input features of each stream to latent features.
The processor exchanges local information on the graph and global information through tokens, and combines the two streams.
The decoder maps the latent features to the residual fields.

\paragraph{Encoder}
The node and element features of each stream are encoded by separate MLPs, and a learnable type embedding is added to distinguish nodes from elements.
This gives the latent node and element features $\bm{H}_{\mathrm{N}}^{\mathrm{B}}$ and $\bm{H}_{\mathrm{E}}^{\mathrm{B}}$ of the baseline stream, and $\bm{H}_{\mathrm{N}}^{\mathrm{P}}$ and $\bm{H}_{\mathrm{E}}^{\mathrm{P}}$ of the perturbation stream.

\paragraph{Processor}
The processor consists of four parts: local message passing in each stream, a token-based global module, a gated fusion of the two streams, and a second local message passing step.

First, local information is exchanged between neighbouring nodes and elements within each stream.
In each stream, one bidirectional GraphSAGE layer \cite{Zhao2026BiLoGo} is applied.
Each element is first updated with the mean of the features of its nodes, and each node is then updated with the mean of the features of its connected elements.
This gives the updated features $\bm{H}_{\mathrm{N}}^{\mathrm{B}*}$, $\bm{H}_{\mathrm{E}}^{\mathrm{B}*}$, $\bm{H}_{\mathrm{N}}^{\mathrm{P}*}$ and $\bm{H}_{\mathrm{E}}^{\mathrm{P}*}$.
The two streams are kept separate at this stage, so that the local perturbation is not averaged with the baseline state.

Second, the token-based global module computes the interaction between distant regions and between the two streams on a small number of tokens.
In each stream, the updated node and element features are combined into one entity set,
\begin{equation}
\bm{H}^{s} = \left[\bm{H}_{\mathrm{N}}^{s*};\, \bm{H}_{\mathrm{E}}^{s*}\right], \quad s \in \{\mathrm{B}, \mathrm{P}\},
\label{eq:res_entity_set}
\end{equation}
where $[\,\cdot\,;\,\cdot\,]$ denotes concatenation along the entity dimension.
Following the slice-based tokeniser of Transolver \cite{Wu2024Transolver,Zhao2026BiLoGo}, a linear layer with a softmax function assigns each entity to 16 slices with soft weights.
Each token is the weighted mean of the entity features in its slice.
This gives the assignment weights $\bm{\Pi}^{s}$ and the tokens $\bm{Z}^{s}$ of each stream.
The two streams use separate slicers, so that the slices of the perturbation stream can follow the regions affected by the ribs, as shown in Fig.~\ref{fig:res_slices}.
Since the attention is computed on 16 tokens instead of all nodes and elements, distant regions of the part can interact at a low cost.
The tokens of each stream are first updated by one Transformer block with self-attention, which gives $\bar{\bm{Z}}^{\mathrm{B}}$ and $\bar{\bm{Z}}^{\mathrm{P}}$.
The perturbation tokens then attend to the baseline tokens through cross-attention,
\begin{equation}
\bm{Z}^{\mathrm{C}} = \bar{\bm{Z}}^{\mathrm{P}} + \mathrm{CrossAttn}\left(\bar{\bm{Z}}^{\mathrm{P}}, \bar{\bm{Z}}^{\mathrm{B}}, \bar{\bm{Z}}^{\mathrm{B}}\right),
\label{eq:res_cross}
\end{equation}
where the three arguments of $\mathrm{CrossAttn}$ are the queries, keys and values, respectively, and $\bm{Z}^{\mathrm{C}}$ denotes the cross-attended tokens.
The cross-attention is followed by a feed-forward layer with a residual connection.
The perturbation tokens are used as queries because the output is the change caused by the ribs.
In this way, each perturbation token collects the baseline state that is relevant to its region.
The cross-attention is one-way, and the baseline tokens are not updated by the perturbation tokens.

The cross-attended tokens are then broadcast back to the entities with the assignment weights of the perturbation stream,
\begin{equation}
\bm{H}^{\mathrm{R}} = \bm{\Pi}^{\mathrm{P}} \bm{Z}^{\mathrm{C}},
\label{eq:res_broadcast}
\end{equation}
and $\bm{H}^{\mathrm{R}}$ is split into the node part $\bm{H}_{\mathrm{N}}^{\mathrm{R}}$ and the element part $\bm{H}_{\mathrm{E}}^{\mathrm{R}}$.

Third, the entity-level baseline features are added again through a gated fusion.
Since the entities in one slice receive similar features, the token-level update may smooth the baseline state within each slice.
The gated fusion is

\begin{equation}
\bm{H}_{q}^{\mathrm{F}} = \bm{H}_{q}^{\mathrm{R}} + \alpha_{q}\, \bm{g}_{q}\!\left(\bm{H}_{q}^{\mathrm{R}}, \bm{H}_{q}^{\mathrm{B}*}\right) \odot \phi_{q}\!\left(\bm{H}_{q}^{\mathrm{B}*}\right), \quad q \in \{\mathrm{N}, \mathrm{E}\},
\label{eq:res_gate}
\end{equation}
where $\phi_{q}$ is an MLP that projects the baseline features, and $\odot$ denotes element-wise multiplication.
The gate $\bm{g}_{q}$ is computed by an MLP with a sigmoid output from the broadcast and baseline features, and it gives a value between 0 and 1 for each feature channel of each entity.
Therefore, the gate controls how much of each channel of the baseline state is added to the broadcast features.
The scalar $\alpha_{q}$ is learnable and is initialised to zero, so that the training starts from the broadcast features and the baseline features are added gradually.

Finally, a second bidirectional GraphSAGE layer is applied to $\bm{H}_{\mathrm{N}}^{\mathrm{F}}$ and $\bm{H}_{\mathrm{E}}^{\mathrm{F}}$, which gives $\bm{H}_{\mathrm{N}}^{\mathrm{post}}$ and $\bm{H}_{\mathrm{E}}^{\mathrm{post}}$.

\paragraph{Decoder}
The node and element decoders are MLPs.
Their inputs are the refined features together with the baseline and perturbation features of the same entity, which are passed directly from the encoder and the first message passing step (dashed lines in Fig.~\ref{fig:res_architecture}).
The decoders give the residual displacement $\Delta\hat{\bm{u}}_i$ and the residual thinning $\Delta\hat{\theta}_j$, respectively.

All latent features have a dimension of 128, and all attention layers use 4 heads.

\section{Baseline comparison and ablation study}
\label{sec:res_source_model}

This section compares R-BiLoGo with two baseline models and examines its components with an ablation study.
First, R-BiLoGo is compared with a Transolver-based model and a hierarchical graph-based model, which are trained on the same residual targets.
The comparison is made on the source dataset and after transfer learning to three selected target geometries with only 20 target simulations, to examine whether the proposed architecture is better suited to learning and transferring the residual fields.
Second, an ablation study examines the residual targets, the baseline response as input and the baseline fusion.
Section~\ref{sec:res_training} describes the training and transfer settings, Section~\ref{sec:res_source_accuracy} presents the baseline comparison, and Section~\ref{sec:res_ablation} presents the ablation study.

\subsection{Training and transfer settings}
\label{sec:res_training}

The surrogate was trained on the source dataset with the loss
\begin{equation}
L = \frac{\mathrm{MSE}\left(\Delta\hat{\bm{u}}, \Delta\bm{u}\right)}{\mathrm{RMS}_{\mathrm{N}}^{2}} + \frac{\mathrm{MSE}\left(\Delta\hat{\theta}, \Delta\theta\right)}{\mathrm{RMS}_{\mathrm{E}}^{2}},
\label{eq:res_loss}
\end{equation}
where $\mathrm{MSE}$ denotes the mean squared error over all nodes or elements, $\mathrm{RMS}_{\mathrm{N}}$ and $\mathrm{RMS}_{\mathrm{E}}$ denote the root mean square values of the residual displacement and residual thinning in the source training data.
The two terms are normalised so that the node and element outputs contribute on a similar scale.
The source model was trained for 1,500 epochs with the AdamW optimiser, a constant learning rate of $1 \times 10^{-3}$, a weight decay of $1 \times 10^{-5}$ and a batch size of 4.

The evaluation metrics follow \cite{Zhao2026BiLoGo}.
Each metric is computed for each test simulation and then averaged over the 50 test simulations.
The mean positional Euclidean error, $e_{\mathrm{N}}$, is the mean Euclidean distance between the predicted and FE final nodal positions of the component with ribs.
Since the baseline response is taken from FE, $e_{\mathrm{N}}$ is also the mean Euclidean error of the residual displacement.
The threshold-based thinning error, $e_{\mathrm{E}}$, is the mean absolute thinning error over the elements whose FE thinning magnitude exceeds 0.005, so that the regions with almost no deformation are excluded.
Since the baseline response is taken from FE, the thinning error of each element is also the error of its residual thinning.
The top-$p$\% critical relative thinning/thickening error, $e_{\mathrm{E}}^{\mathrm{top}}$, is computed at both ends of the thinning distribution.
The first value is the mean relative thinning error over the $p$\% of elements with the largest FE thinning, and the second value is that over the $p$\% of elements with the largest FE thickening.
$e_{\mathrm{E}}^{\mathrm{top}}$ is the average of the two values.
Following \cite{Zhao2026BiLoGo} for the same mesh, $p = 0.25$ is used.

R-BiLoGo is compared with two baseline models, which predict the same residual targets.
The first baseline, R-Transolver, is a vanilla Transolver \cite{Wu2024Transolver} with the same setup as R-BiLoGo, but without the node--element message passing and the separate baseline and perturbation streams.
The second baseline, R-CAtt-BiUGNN, is a hierarchical bipartite graph neural network \cite{Zhao2026CAttBiGNN}.
Each baseline model used the input representation and target normalisation of its original implementation, while R-BiLoGo used the inputs in Table~\ref{tab:res_features} and the loss in Eq.~\eqref{eq:res_loss}.
All models were trained on the 193 source training simulations and evaluated on the 50 source test simulations.

\paragraph{Transfer settings}
\label{sec:res_transfer_protocol}

For the three selected target geometries, the number of target training simulations was $K \in \{5, 10, 20, 40, 80, 160\}$.
The nested subsets of Eq.~\eqref{eq:res_subsets} were used.
In addition, the source model was applied to the target test set without transfer learning, which is referred to as zero-shot prediction.
The model trained from scratch on all simulations of the training pool is referred to as the oracle, and is used as a reference for the lowest error that can be obtained with the available target data.
R-Transolver and R-CAtt-BiUGNN were also trained on the source dataset and transferred with the same protocol, so that the transfer of the three models can be compared.

For each target geometry and each $K$, two models with the same architecture were trained.
The first model was initialised from the source model and then trained on the target training simulations.
The second model was trained from random initialisation on the same target training simulations.
For the 22 target geometries in Section~\ref{sec:res_systematic}, two target-data budgets were used, with $K = 20$ and $K = 80$ training simulations, which are referred to as K20 and K80.
K20 and K80 use the subsets $S_{20}$ and $S_{80}$, respectively.
The K20 and K80 models were trained in separate runs, and each transfer-learning run started from the source model.
Each setting was trained once.
All models of a target geometry use the same validation and test simulations.

All parameters were updated during transfer learning, and no module was frozen or reinitialised.
The parameters were loaded from the source model, while the optimiser state was reinitialised.
Transfer learning used the AdamW optimiser with a learning rate of $1 \times 10^{-4}$, a weight decay of $1 \times 10^{-5}$, a batch size of 4 and gradient clipping at a global norm of 1.0, without a learning-rate scheduler.
Each run was trained for 20,000 optimiser steps, which corresponds to 4,000 epochs for K20 and 1,000 epochs for K80.
The same number of steps was used for both budgets, so that the two budgets receive the same number of parameter updates.
Training from scratch used the same number of steps and the same settings, except that the learning rate was $1 \times 10^{-3}$, which is the learning rate used to train the source model.
The learning rate of transfer learning was set ten times smaller than that of training from scratch, so that transfer learning does not move the parameters far from the source model at the start of training.
The validation loss on the validation simulations, defined in the same form as Eq.~\eqref{eq:res_loss}, was evaluated every 250 steps.
The model with the lowest validation loss was used for testing.
The runs for the 22 target geometries were performed on one NVIDIA L4 GPU. Each run took approximately 2.4~h.
Therefore, transfer does not reduce the training time in this protocol, and its benefit is evaluated in terms of prediction accuracy and the number of target simulations.

The transfer benefit is measured by the relative transfer gain for each metric,
\begin{equation}
G = \frac{e^{\mathrm{scratch}} - e^{\mathrm{TL}}}{e^{\mathrm{scratch}}} \times 100\%,
\label{eq:res_gain}
\end{equation}
where $e^{\mathrm{scratch}}$ and $e^{\mathrm{TL}}$ denote the test errors of the model trained from scratch and the transferred model, respectively, with the same target-data budget.
A positive $G$ indicates that transfer reduces the error, and a negative $G$ indicates negative transfer.

The three types of comparison are used for different purposes.
First, the test errors after transfer learning with the same $K$ are used to compare the models, since they show which model gives the most accurate prediction for the same number of target FE simulations.
The gain is not used here, since a model with a high error when trained from scratch can show a large gain.
Second, the gain is used to compare the target geometries, since it is computed relative to training from scratch on the same target geometry, which reduces the effect of the difficulty of each target.
Third, the comparison between the transferred K20 model and the K80 model trained from scratch is used to estimate the reduction in target simulations, which corresponds to the reduction in FE cost discussed in Section~\ref{sec:res_efficiency}.

\subsection{Baseline comparison}
\label{sec:res_source_accuracy}

R-BiLoGo is compared with the two baseline models first on the source dataset and then after transfer learning with 20 target simulations.

\paragraph{Accuracy on the source dataset}

\begin{figure}[!htbp]
\centering
\includegraphics[width=\textwidth]{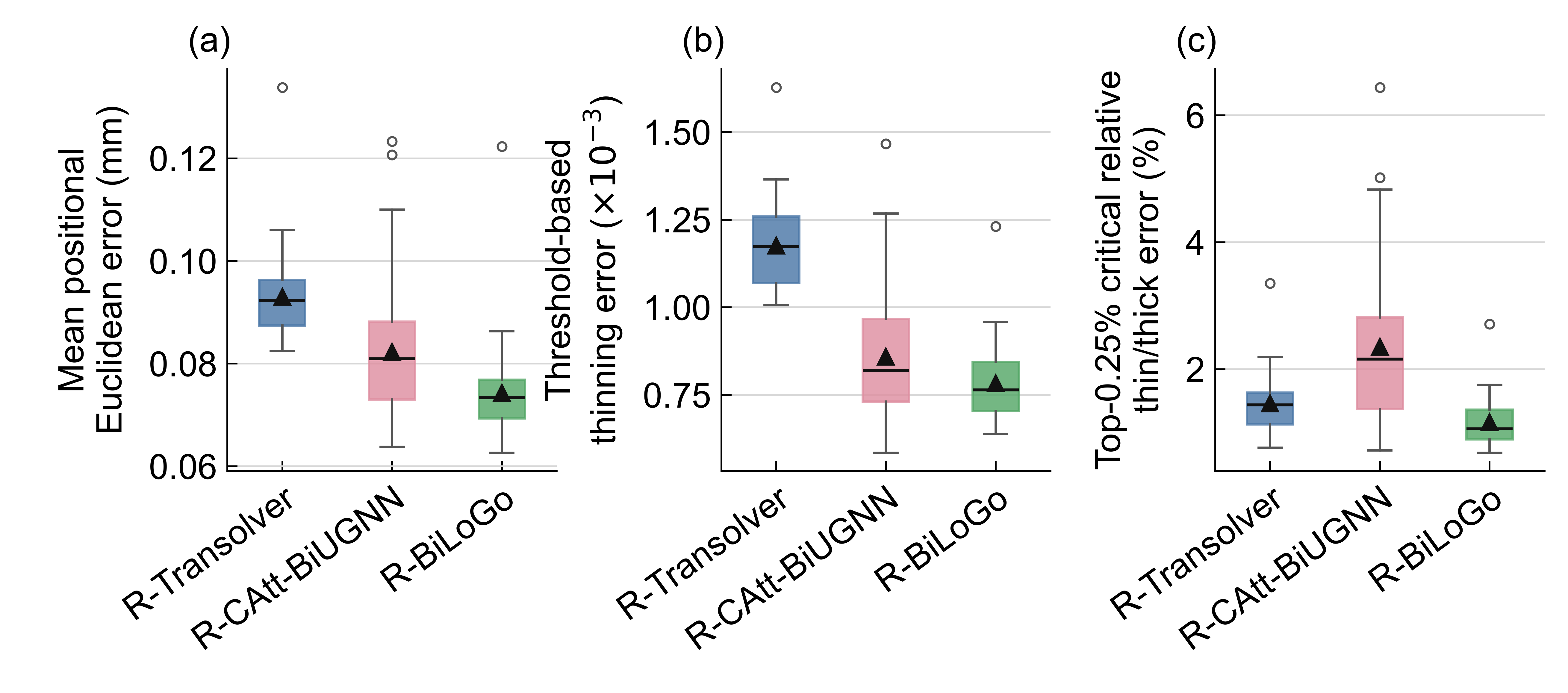}
\caption{Distributions of the errors over the 50 source test simulations for R-Transolver, R-CAtt-BiUGNN and R-BiLoGo. (a) Mean positional Euclidean error. (b) Threshold-based thinning error. (c) Top-0.25\% critical relative thinning/thickening error. The triangles show the mean values.}
\label{fig:res_source_box}
\end{figure}

The accuracy and the number of parameters of the three models are listed in Table~\ref{tab:res_baselines}, and the distributions of the errors over the test simulations are shown in Fig.~\ref{fig:res_source_box}.

\begin{table}[!htbp]
\centering
\caption{Accuracy and number of parameters of R-BiLoGo and the baseline models on the source test set. The best value in each column is shown in bold and the second best is underlined.}
\label{tab:res_baselines}
\footnotesize
\setlength{\tabcolsep}{4pt}
\begin{tabular}{@{}lcccc@{}}
\toprule
Model & $e_{\mathrm{N}}$ (mm) & $e_{\mathrm{E}}$ ($\times 10^{-4}$) & $e_{\mathrm{E}}^{\mathrm{top}}$ (\%) & Parameters \\
\midrule
R-Transolver & 0.0930 & 11.75 & \underline{1.46} & \textbf{0.31M} \\
R-CAtt-BiUGNN & \underline{0.0822} & \underline{8.58} & 2.35 & 939.9M \\
R-BiLoGo & \textbf{0.0743} & \textbf{7.82} & \textbf{1.16} & \underline{1.48M} \\
\bottomrule
\end{tabular}
\end{table}

R-BiLoGo gave the lowest mean errors and the lowest median errors for all three metrics.
Compared with R-CAtt-BiUGNN, the median thinning errors were similar, but the errors of R-CAtt-BiUGNN varied more between the test simulations, especially for $e_{\mathrm{E}}^{\mathrm{top}}$.
This indicates that R-BiLoGo predicted the critical thinning and thickening more consistently.
R-Transolver has the fewest parameters, while R-CAtt-BiUGNN has a larger number of parameters than the other two models.
This is mainly due to its inter-level mappings, which depend on the graph levels of the mesh \cite{Zhao2026BiLoGo}.
In contrast, the number of parameters of R-BiLoGo does not depend on the mesh size.

\begin{figure}[!htbp]
\centering
\includegraphics[width=\textwidth]{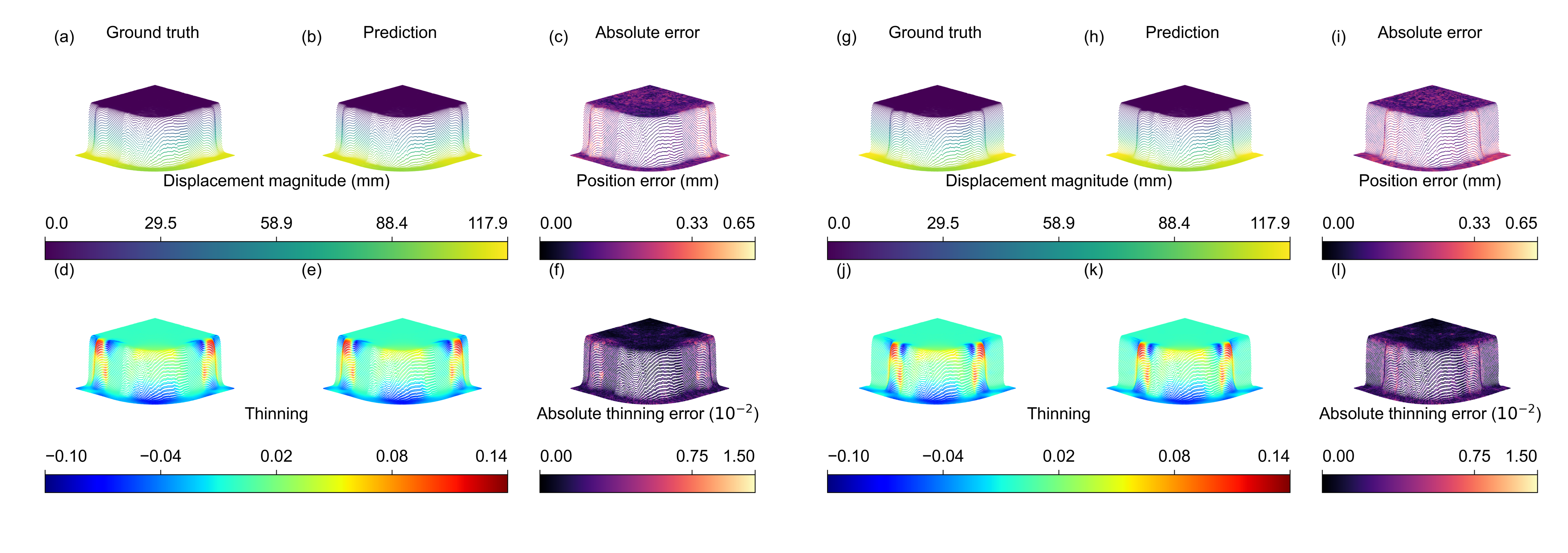}
\caption{Complete fields predicted by R-BiLoGo for two source test simulations. (a)--(f) First test simulation. (g)--(l) Second test simulation. (a), (g) FE displacement magnitude. (b), (h) Predicted displacement magnitude. (c), (i) Position error. (d), (j) FE thinning. (e), (k) Predicted thinning. (f), (l) Absolute thinning error. The predicted fields were obtained by adding the predicted residuals to the baseline response.}
\label{fig:res_source_fields}
\end{figure}

Fig.~\ref{fig:res_source_fields} shows the complete fields predicted by R-BiLoGo for two source test simulations.
The predicted fields were obtained by adding the predicted residual displacement and residual thinning to the baseline response.
For both simulations, the predicted displacement and thinning distributions agreed with the FE results.
The position errors were small compared with the displacement magnitude, and the larger errors were mainly located on the side walls and near the corners.
The high-thinning regions on the side walls near the corners were reproduced in both location and magnitude.
The absolute thinning errors were larger on the side walls than on the flat region of the component.

\begin{figure}[!htbp]
\centering
\includegraphics[width=0.85\textwidth]{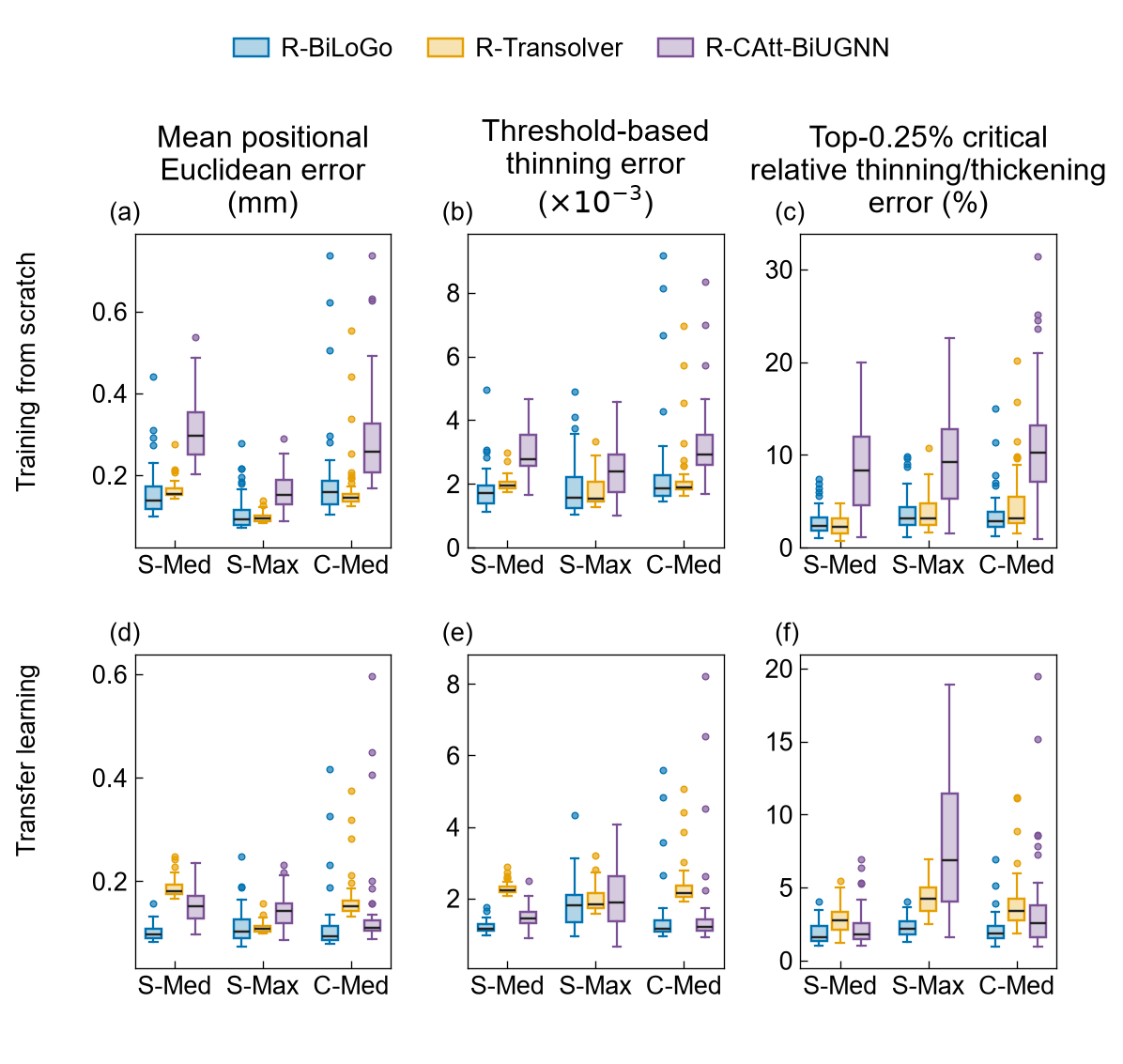}
\caption{Distributions of the errors of R-BiLoGo, R-Transolver and R-CAtt-BiUGNN over the 50 test simulations of S-Med, S-Max and C-Med with $K = 20$. (a)--(c) Training from scratch. (d)--(f) Transfer learning. (a), (d) Mean positional Euclidean error. (b), (e) Threshold-based thinning error. (c), (f) Top-0.25\% critical relative thinning/thickening error.}
\label{fig:res_k20}
\end{figure}

\paragraph{Accuracy after transfer learning with 20 target simulations}
\label{sec:res_selected}
The three models were transferred to S-Med, S-Max and C-Med with 20 target simulations, which is about one tenth of the 193 source training simulations.
The distributions of the errors are shown in Fig.~\ref{fig:res_k20}, and the mean errors, including those of the zero-shot prediction, are listed in Table~\ref{tab:res_selected_k20}.

\begin{table}[!htbp]
\centering
\caption{Mean test errors on the three selected target geometries with $K = 20$. ZS denotes zero-shot prediction, TL denotes transfer learning and S denotes training from scratch. For each target geometry, the best value in each column is shown in bold and the second best is underlined.}
\label{tab:res_selected_k20}
\footnotesize
\setlength{\tabcolsep}{3pt}
\begin{tabular}{@{}llccccccccc@{}}
\toprule
Target & Model & \multicolumn{3}{c}{$e_{\mathrm{N}}$ (mm)} & \multicolumn{3}{c}{$e_{\mathrm{E}}$ ($\times 10^{-3}$)} & \multicolumn{3}{c}{$e_{\mathrm{E}}^{\mathrm{top}}$ (\%)} \\
\cmidrule(lr){3-5}\cmidrule(lr){6-8}\cmidrule(l){9-11}
 & & ZS & TL & S & ZS & TL & S & ZS & TL & S \\
\midrule
S-Med & R-Transolver & 2.1960 & 0.1868 & \underline{0.1632} & 50.88 & 2.27 & \underline{1.99} & 60.96 & 2.82 & \textbf{2.46} \\
& R-CAtt-BiUGNN & \textbf{0.7039} & \underline{0.1532} & 0.3159 & \textbf{5.94} & \underline{1.47} & 3.01 & \textbf{8.91} & \underline{2.33} & 8.52 \\
& R-BiLoGo & \underline{1.7410} & \textbf{0.1005} & \textbf{0.1595} & \underline{16.41} & \textbf{1.20} & \textbf{1.81} & \underline{29.21} & \textbf{1.93} & \underline{2.76} \\
\midrule
S-Max & R-Transolver & 2.8866 & \textbf{0.1107} & \textbf{0.0972} & 70.94 & \underline{1.97} & \textbf{1.75} & 193.37 & \underline{4.26} & \underline{3.84} \\
& R-CAtt-BiUGNN & \underline{1.2070} & 0.1456 & 0.1608 & \textbf{10.70} & 2.04 & 2.42 & \underline{40.23} & 8.28 & 9.44 \\
& R-BiLoGo & \textbf{0.9709} & \underline{0.1128} & \underline{0.1090} & \underline{13.53} & \textbf{1.84} & \underline{1.88} & \textbf{18.92} & \textbf{2.25} & \textbf{3.80} \\
\midrule
C-Med & R-Transolver & 2.4760 & 0.1652 & \textbf{0.1673} & 50.76 & 2.32 & \textbf{2.20} & 85.62 & 3.86 & \underline{4.82} \\
& R-CAtt-BiUGNN & \textbf{0.3193} & \underline{0.1376} & 0.2881 & \textbf{3.67} & \underline{1.57} & 3.22 & \textbf{8.42} & \underline{3.49} & 11.15 \\
& R-BiLoGo & \underline{1.4699} & \textbf{0.1121} & \underline{0.1886} & \underline{15.27} & \textbf{1.44} & \underline{2.36} & \underline{28.49} & \textbf{2.09} & \textbf{3.55} \\
\bottomrule
\end{tabular}
\end{table}

The zero-shot errors of all three models were much higher than the errors after transfer learning and after training from scratch.
This indicates a clear distribution shift between the source and target datasets, which agrees with Section~\ref{sec:res_formulation}, in which the effect of the ribs was expected to depend on the global geometry.
Therefore, target simulations were required to adapt the source model.

With 20 target simulations, R-BiLoGo and R-CAtt-BiUGNN showed positive transfer on S-Med and C-Med, with lower errors after transfer learning than after training from scratch.
R-CAtt-BiUGNN also showed positive transfer on S-Max.
On S-Max, which has the largest change in global geometry among the three targets, R-BiLoGo gave a slightly higher node error after transfer learning than after training from scratch, but a lower critical relative thinning/thickening error.
Among the three models, R-BiLoGo gave the lowest errors after transfer learning overall.
The large transfer gains of R-CAtt-BiUGNN were mainly caused by its high errors when trained from scratch.
In contrast, R-Transolver showed little or negative transfer in most cases.

The computational cost of transfer learning is listed in Table~\ref{tab:res_ft_cost}.
R-Transolver required the least training time and GPU memory, since it has the fewest parameters, while R-CAtt-BiUGNN required the most.

\begin{table}[!htbp]
\centering
\caption{Computational cost of transfer learning on S-Max with $K = 20$, measured on an NVIDIA A40 GPU. One epoch is one pass over the 20 target training simulations. The batch size was 4 for R-Transolver and R-BiLoGo and 1 for R-CAtt-BiUGNN. The best value in each column is shown in bold and the second best is underlined.}
\label{tab:res_ft_cost}
\small
\begin{tabular}{@{}lcc@{}}
\toprule
Model & Training time per epoch (s) & Peak GPU memory (GiB) \\
\midrule
R-Transolver & \textbf{0.46} & \textbf{6.7} \\
R-CAtt-BiUGNN & 5.93 & 15.3 \\
R-BiLoGo & \underline{1.25} & \underline{11.6} \\
\bottomrule
\end{tabular}
\end{table}

Considering both the accuracy and the computational cost, R-BiLoGo is used in the rest of the study.
R-CAtt-BiUGNN required the most training time and GPU memory, and R-Transolver showed negative transfer in most cases.
In Section~\ref{sec:res_transfer}, R-Transolver is still compared with R-BiLoGo for different numbers of target simulations, since its cost is the lowest.

\subsection{Ablation study}
\label{sec:res_ablation}

The ablation study examines the two parts of the baseline-conditioned residual formulation and the baseline fusion of R-BiLoGo.
First, the residual targets were replaced by the complete fields.
Second, the baseline response was removed from the input features.
Third, the baseline fusion of Eq.~\eqref{eq:res_gate} was removed.
In each case, only one part was changed, and the other settings were the same as those of R-BiLoGo.
The results on the source test set are listed in Table~\ref{tab:res_ablation}.

\begin{table}[!htbp]
\centering
\caption{Ablations of R-BiLoGo on the source test set. The best value in each column is shown in bold and the second best is underlined.}
\label{tab:res_ablation}
\small
\begin{tabular}{@{}lccc@{}}
\toprule
Model & $e_{\mathrm{N}}$ (mm) & $e_{\mathrm{E}}$ ($\times 10^{-4}$) & $e_{\mathrm{E}}^{\mathrm{top}}$ (\%) \\
\midrule
R-BiLoGo & \textbf{0.0743} & \textbf{7.82} & \underline{1.16} \\
Complete-field targets & 0.1548 & \underline{7.92} & \textbf{1.13} \\
w/o baseline response & 0.0783 & 8.18 & 1.18 \\
w/o baseline fusion & \underline{0.0775} & 10.3 & 1.41 \\
\bottomrule
\end{tabular}
\end{table}

\paragraph{Residual and complete-field targets}
To examine the formulation, a model with the same architecture and the same input features as R-BiLoGo was trained to predict the complete final positions and thinning of the component with ribs instead of the residual fields.
This model is referred to as the complete-field model.
Compared with R-BiLoGo, the complete-field model increased $e_{\mathrm{N}}$ from 0.0743~mm to 0.1548~mm, which is an increase of 108.4\%.
In contrast, its $e_{\mathrm{E}}$ was $7.92 \times 10^{-4}$, which is only 1.3\% higher than that of R-BiLoGo.
Its $e_{\mathrm{E}}^{\mathrm{top}}$ was 1.13\%, which is close to that of R-BiLoGo, 1.16\%.
Therefore, the residual formulation mainly reduced the node error, while the two formulations gave similar thinning errors.
This may be attributed to the different magnitudes of the complete and residual displacements.
The complete displacement reaches about 118~mm (Fig.~\ref{fig:res_source_fields}), while the change caused by the ribs is much smaller.
When the complete positions are predicted, a small relative error in the global deformation can be large compared with the effect of the ribs.
In contrast, with the residual targets, the global deformation is taken from the baseline FE result, and the model only predicts the change caused by the ribs.
For the thinning, the baseline thinning is an input of both models, and the complete thinning does not contain a large global component.
This may explain why the two formulations gave similar thinning errors.
The complete-field model was also transferred to S-Med and C-Med with the same 20 target simulations and the same settings as in Section~\ref{sec:res_selected}.
After transfer learning, its node errors were 0.1950~mm for S-Med and 0.2106~mm for C-Med, which are 94.1\% and 87.9\% higher than those of R-BiLoGo.
Therefore, the effect of the formulation on the source dataset remained after transfer.
Compared with training from scratch, transfer learning reduced the node error of the complete-field model by 37.5\% for S-Med and 37.7\% for C-Med.
These reductions are similar to those of R-BiLoGo, which indicates that the transfer gain was similar for the two formulations.
However, the complete-field model after transfer learning still gave higher node errors than R-BiLoGo trained from scratch on the same 20 simulations, which were 0.1595~mm for S-Med and 0.1886~mm for C-Med.

\paragraph{Baseline response as input}
To examine the role of the baseline response, the baseline positions and the baseline thinning were removed from the input features, while the targets were still the residual fields.
The errors on S-Med with 20 target simulations are listed in Table~\ref{tab:res_ablation_smed}.

\begin{table}[!htbp]
\centering
\caption{Errors of R-BiLoGo and the model without the baseline response on S-Med with $K = 20$. ZS denotes zero-shot prediction, TL denotes transfer learning and S denotes training from scratch. The better value in each column is shown in bold.}
\label{tab:res_ablation_smed}
\footnotesize
\setlength{\tabcolsep}{3pt}
\begin{tabular}{@{}lccccccccc@{}}
\toprule
Model & \multicolumn{3}{c}{$e_{\mathrm{N}}$ (mm)} & \multicolumn{3}{c}{$e_{\mathrm{E}}$ ($\times 10^{-3}$)} & \multicolumn{3}{c}{$e_{\mathrm{E}}^{\mathrm{top}}$ (\%)} \\
\cmidrule(lr){2-4}\cmidrule(lr){5-7}\cmidrule(l){8-10}
 & ZS & TL & S & ZS & TL & S & ZS & TL & S \\
\midrule
R-BiLoGo & \textbf{1.7410} & \textbf{0.1005} & \textbf{0.1595} & \textbf{16.41} & \textbf{1.20} & \textbf{1.81} & \textbf{29.21} & \textbf{1.93} & \textbf{2.76} \\
w/o baseline response & 1.7852 & 0.1020 & 0.1778 & 25.21 & 1.30 & 2.00 & 81.17 & 2.06 & 3.57 \\
\bottomrule
\end{tabular}
\end{table}

Without the baseline response, all errors increased, both on the source test set and on S-Med.
On the source test set, $e_{\mathrm{N}}$ and $e_{\mathrm{E}}$ increased by 5.4\% and 4.6\%.
After transfer learning to S-Med, $e_{\mathrm{E}}$ and $e_{\mathrm{E}}^{\mathrm{top}}$ increased by 8.7\% and 6.9\%, while $e_{\mathrm{N}}$ increased by only 1.5\%.
The effect was larger when few or no target simulations were used.
Training from scratch with 20 target simulations gave 10.3\% and 29\% higher $e_{\mathrm{E}}$ and $e_{\mathrm{E}}^{\mathrm{top}}$, and the zero-shot $e_{\mathrm{E}}$ increased by 54\%.
Therefore, the baseline response improved the accuracy in all settings, mainly for the thinning.
It should be noted that, in this study, the source model and each transferred model were trained on the simulations of one global geometry, so all simulations in each training set share the same baseline response.
Therefore, the importance of the baseline response and the gated baseline fusion may not be fully shown here.
When one model is trained on a large and diverse set of global geometries, the baseline response may help the model learn how the effect of the local features changes with the global geometry.

\paragraph{Baseline fusion}
Without the baseline fusion, $e_{\mathrm{N}}$ increased by 4.3\%, while $e_{\mathrm{E}}$ increased by 31.7\% and $e_{\mathrm{E}}^{\mathrm{top}}$ by 21.6\% compared with R-BiLoGo.
Therefore, the baseline fusion mainly reduced the thinning error.
This may be attributed to the element-level baseline features, which are added after the token-level update and keep the local baseline state of each element.

\begin{figure}[!htbp]
\centering
\includegraphics[width=\textwidth]{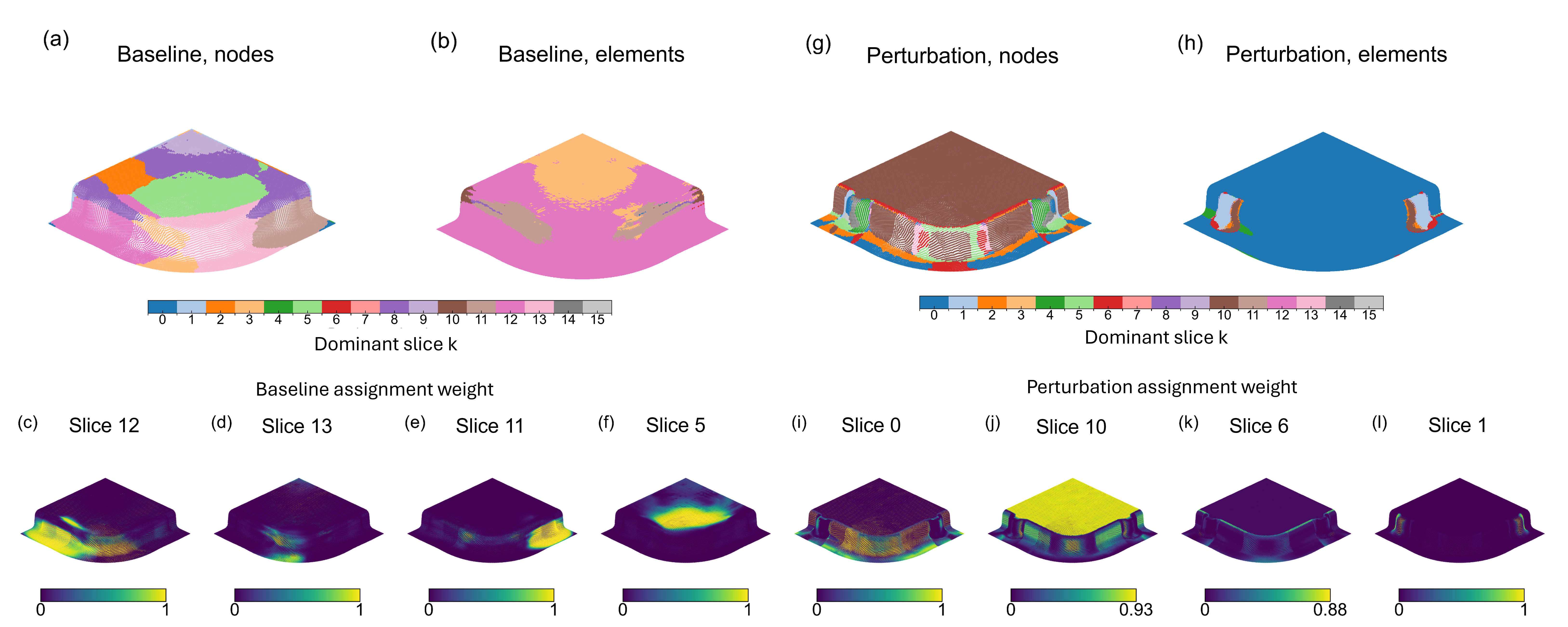}
\caption{Slice assignment of R-BiLoGo for an example test simulation of the source dataset. (a)--(f) Baseline stream. (g)--(l) Perturbation stream. (a), (g) Dominant slice of each node. (b), (h) Dominant slice of each element. (c)--(f) Assignment weights of baseline slices 12, 13, 11 and 5. (i)--(l) Assignment weights of perturbation slices 0, 10, 6 and 1. Each weight panel uses its own colour scale.}
\label{fig:res_slices}
\end{figure}

\paragraph{Slice assignment}
The slice assignment of R-BiLoGo is shown in Fig.~\ref{fig:res_slices} for an example test simulation of the source dataset.
In the baseline stream, the dominant slices of the nodes form large connected regions, which divide the part into the flange, the side wall, the corner and the flat region of the part (Fig.~\ref{fig:res_slices}(a)).
In the perturbation stream, most elements are assigned to one slice, and only the elements near the ribs are assigned to other slices (Fig.~\ref{fig:res_slices}(h)).
The assignment weights of perturbation slice 1 are concentrated at the ribs (Fig.~\ref{fig:res_slices}(l)), and those of perturbation slice 6 are concentrated along the upper and lower edges of the side wall (Fig.~\ref{fig:res_slices}(k)).
Therefore, the perturbation stream separates the rib regions from the rest of the part, while the baseline stream divides the part by region.
This may be attributed to the residual formulation, in which the response away from the ribs is mainly described by the baseline.

\section{Effect of the number of target simulations on transfer learning}
\label{sec:res_transfer}
\label{sec:res_ksweep}

\begin{figure}[!htbp]
\centering
\includegraphics[width=\textwidth]{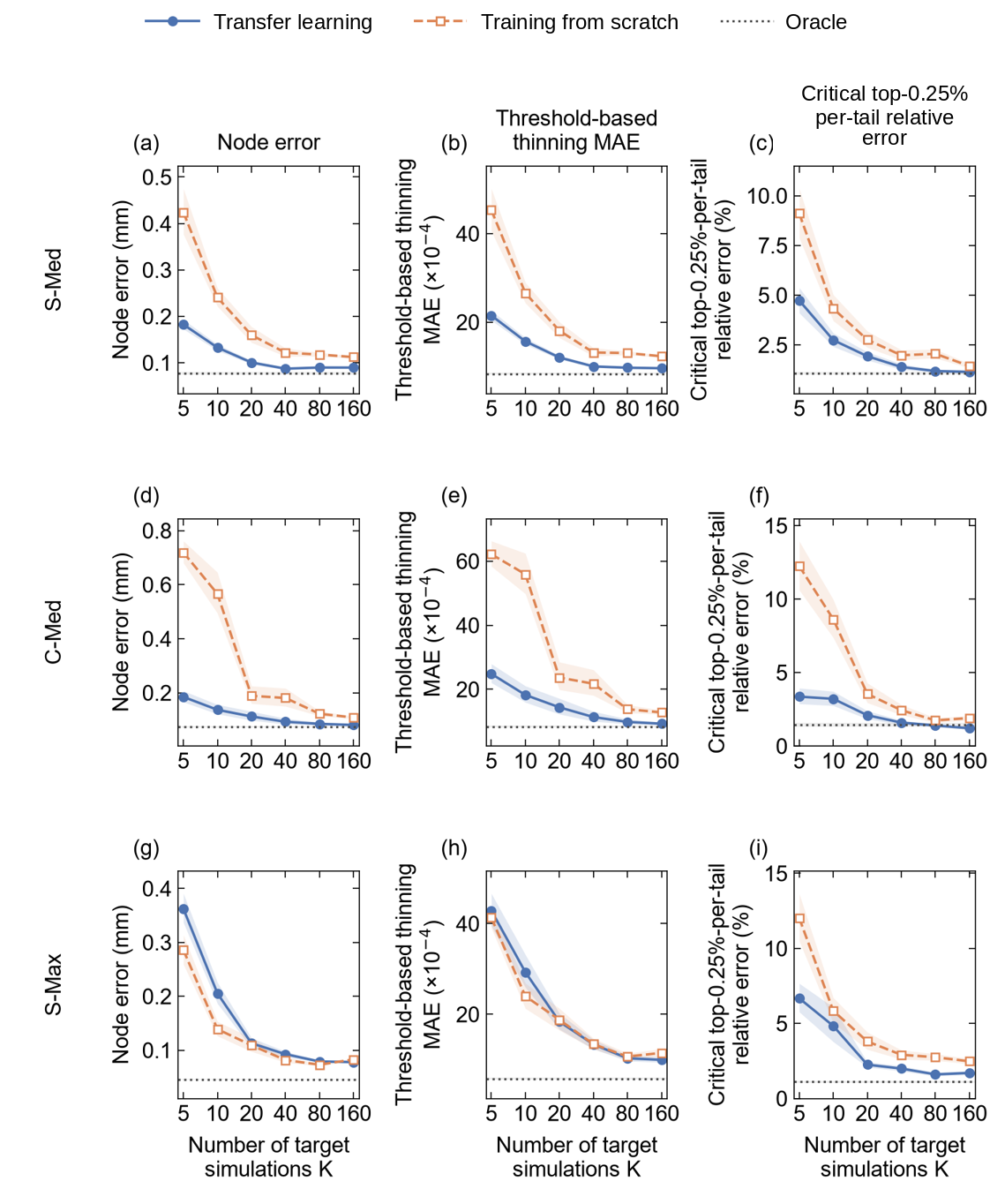}
\caption{Test errors of R-BiLoGo against the number of target simulations $K$ for transfer learning (solid lines) and training from scratch (dashed lines). The shaded bands show the standard deviation of the errors over the 50 test simulations. The dotted lines show the errors of the oracle, which was trained from scratch on the full training pool of 294 simulations. Rows: S-Med, C-Med and S-Max. Columns: mean positional Euclidean error, threshold-based thinning error and top-0.25\% critical relative thinning/thickening error.}
\label{fig:res_learning_curves}
\end{figure}

This section examines the effect of the number of target simulations $K$ on transfer learning.
Transfer learning and training from scratch were compared for $K$ from 5 to 160 on three target geometries.
For each $K$, the target training simulations were taken from the nested subsets of Eq.~\eqref{eq:res_subsets}, which were selected from the training pool as described in Appendix~\ref{app:res_subsets}.
S-Med and S-Max have the median and the maximum Chamfer distance to the source geometry in the Standard subclass.
C-Med has the median Chamfer distance in the Chamfered subclass.
The test errors are shown in Fig.~\ref{fig:res_learning_curves}.
The oracle in Fig.~\ref{fig:res_learning_curves} was trained from scratch on the full training pool of each target geometry, which contains 294 simulations.

The errors of both methods became more stable as $K$ increased.
For $K \le 10$, the errors changed rapidly with $K$.
In particular, training from scratch gave large errors with $K = 5$ and $K = 10$ for S-Med and C-Med.
From $K = 20$, the errors of both methods changed more slowly.
From $K = 80$ to $K = 160$, the errors decreased only slightly.
For S-Med and C-Med, the errors after transfer learning with $K = 80$ approached those of the oracle.

The benefit of transfer learning differed between the target geometries.
For S-Med and C-Med, transfer learning gave lower errors than training from scratch for all values of $K$.
Transfer learning with 20 to 40 target simulations gave lower node errors than training from scratch with 160 simulations.
For S-Max, transfer learning gave higher node errors than training from scratch for $K \le 80$.
Its thinning errors were similar to those of training from scratch for $20 \le K \le 80$, but its critical relative thinning/thickening errors were lower for all values of $K$.
S-Max has the largest Chamfer distance among the three targets.
Therefore, the transfer benefit may depend on the geometric difference between the source and target geometries.
In contrast to R-BiLoGo, the error of the transferred R-Transolver hardly decreased for $K \ge 20$ (Appendix~\ref{app:res_transolver}).

\section{Relationship between geometric difference and transfer benefit}
\label{sec:res_systematic}

In this study, the rib configurations are the same for all target geometries, and only the global geometry changes between the source and the targets.
Therefore, the global geometry is the variable that determines whether the effect of the ribs can be transferred.
Section~\ref{sec:res_transfer} showed that transfer learning did not always give lower errors than training from scratch, and that the transfer benefit differed between the target geometries.
This section therefore examines how the global geometry affects the transfer benefit.
The difference between the source and target global geometries is described by geometric descriptors, and the transfer benefit is related to these descriptors on 22 target geometries.

The study follows three stages, referred to as BUILD, MEASURE and DISCOVER.
BUILD fixes everything except the global geometry, including the model, the transfer-learning settings, the target-data budgets and the evaluation settings.
As a result, the differences in the transfer benefit between the target geometries can be attributed to the global geometry.
MEASURE selects the geometric descriptors that are used in DISCOVER, using geometry information only, before the transfer results are examined.
DISCOVER examines whether and how each selected descriptor affects the transfer benefit, first within the Standard subclass and then across the four subclasses.
The three stages do not depend on the model or the transfer strategy, and they can also be applied to other models and transfer strategies.

Based on the results in Section~\ref{sec:res_transfer}, K20 and K80 were used to examine transfer learning with a small and a large number of target simulations, respectively.
With K80, the errors of both transfer learning and training from scratch were relatively stable.
Therefore, K80 is used to examine whether the transfer is positive or negative, and the transfer gain is computed with K80.
K20 is used as an extreme case with a small number of target simulations.
The transferred model with K20 is compared with the model trained from scratch with K80.
This comparison shows whether transfer learning with 20 target simulations can give errors equal to or lower than those of training from scratch with 80 simulations.
It is used only to estimate how many target simulations can be saved.

\subsection{BUILD: model and transfer protocol}
\label{sec:res_build}

The study uses R-BiLoGo with the source model of Section~\ref{sec:res_source_model}.
R-BiLoGo was selected because it gave the lowest errors after transfer learning in Section~\ref{sec:res_selected} and could make use of additional target simulations in Section~\ref{sec:res_transfer}, while transfer learning of R-Transolver gave little or no benefit compared with training it from scratch.
For each of the 22 target geometries, the model was transferred and trained from scratch with K20 and K80.
The transfer-learning settings, including the optimiser, the learning rate, the number of optimiser steps and the model selection by the validation loss, are the same as in Section~\ref{sec:res_training}.
The evaluation settings are also the same, with 50 test simulations for each target geometry and the three metrics of Section~\ref{sec:res_training}.

\subsection{MEASURE: geometric descriptors}
\label{sec:res_descriptors}

The difference between the source geometry $S$ and a target geometry $T$ is described by descriptors measured from the realised CAD forming surfaces.
In this way, geometries with different parameterisations can be compared with the same descriptors, and the descriptors are therefore referred to as parameter-free.

The descriptors were selected in three steps, using CAD geometry only, before any transfer results were examined (Appendix~\ref{app:res_descriptors}).
First, 28 candidate descriptors were defined to cover five types of geometric difference: surface position, surface orientation, curvature, cross-sectional profile and global dimensions.
Each candidate combines a geometric field, such as the distance between the two surfaces, with a statistic, such as the mean, a percentile or the maximum.
Second, the candidates were checked for a complete definition, stable computation and convergence with the mesh, and 15 candidates remained.
Third, the 15 candidates were computed on 430 CAD geometries of the four subclasses, relative to the source geometry.
A descriptor was retained if it changed clearly within each subclass and was not strongly correlated with the other retained descriptors.
Three descriptors were selected in this way.

Since each descriptor depends on several design parameters in a nonlinear way, a change in only one descriptor cannot be obtained by changing one design parameter.
When suitable geometries could not be found among the measured CAD geometries, subclass-specific radial basis function (RBF) surrogates were fitted where needed to map the CAD parameters to the descriptors.
The surrogates were used to search for CAD parameters that change one descriptor while the changes in the others are limited, and a local null-space search was used to refine this search (Appendix~\ref{app:descriptor_control}).
Each proposed geometry was then generated in CAD and measured again, and the target geometries were selected based on these measured values rather than the surrogate predictions.
A fourth descriptor, the curvedness distribution distance $\delta_c$, was introduced during the DISCOVER stage based on the transfer results, as described in Section~\ref{sec:res_standard}.

The first descriptor is the sampled maximum bidirectional surface distance $\delta_{\max}$, in mm, which measures the largest positional difference between the two surfaces.
The second descriptor is the change in the area-weighted 99th-percentile curvedness $\Delta c_{0.99}^{w}$, in mm$^{-1}$, which measures the change in the high-curvature regions of the surface.
The curvedness is computed from the principal curvatures following \cite{KoenderinkVanDoorn1992}.
The third descriptor is the forming-depth change $\Delta H = H(T) - H(S)$, in mm.
A negative $\Delta H$ indicates that the target is shallower than the source.
The values of the descriptors of the target geometries are shown on the horizontal axes of Figs.~\ref{fig:res_standard_paths} and \ref{fig:res_cross}.

\subsection{DISCOVER: effect of geometric descriptors in the Standard subclass}
\label{sec:res_standard}

This subsection relates the transfer benefit to each descriptor within the Standard subclass.
Since many descriptors change at the same time between two geometries, the effect of one descriptor cannot be identified from arbitrary target geometries.
Therefore, controlled variations were used, in which one descriptor changes while the changes in the others are limited.
The target geometries are named by the subclass (S, C, E and O for Standard, Chamfered, Elliptical and Ogee), the varied descriptor and its level, where $\Delta c$ denotes $\Delta c_{0.99}^{w}$.
For $\delta_{\max}$ and $\delta_c$, the levels are low, mid and high.
For the signed descriptors $\Delta c$ and $\Delta H$, the levels are neg-high, neg-mid, zero, pos-mid and pos-high.
S-$\Delta c$-zero, S-$\Delta H$-zero and S-$\delta_c$-low are the same geometry, and C-$\delta_{\max}$-low and C-$\delta_c$-low are the same geometry.
The relationships were examined with the K80 gain.
With K80, both the transferred model and the model trained from scratch use more target simulations, and the gain is less affected by the selection of the target subset.
In this section, the K80 gain is given as the mean of the relative reductions in the node error and the element error, so that one value describes the transfer benefit of each target geometry.
The node and element gains are also shown separately in the figures.

\begin{figure}[!htbp]
\centering
\includegraphics[width=\textwidth]{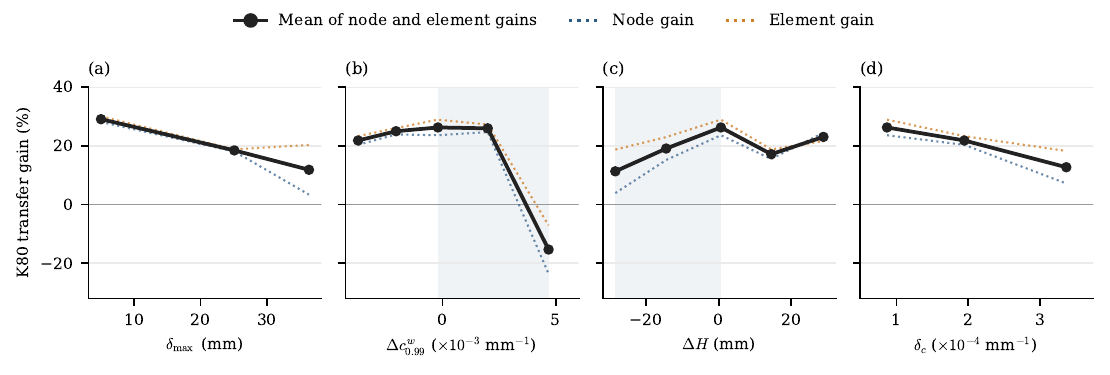}
\caption{K80 transfer gain over the controlled variations of the Standard subclass. (a) $\delta_{\max}$. (b) $\Delta c_{0.99}^{w}$. (c) $\Delta H$. (d) $\delta_c$, for S-$\delta_c$-low, S-$\Delta c$-neg-high and S-$\delta_c$-high. The solid lines show the mean of the node and element gains, and the dotted lines show the node and element gains. The shaded regions show the ranges in which the gain changed.}
\label{fig:res_standard_paths}
\end{figure}

The analysis was performed in two steps.
First, each descriptor was varied over the full range of its controlled variation, in both directions where available (Fig.~\ref{fig:res_standard_paths}).
Second, the range of each descriptor was narrowed to the part in which the gain changed (shaded regions in Fig.~\ref{fig:res_standard_paths}).
Since the ranges were narrowed after the results were examined, the trends are checked on the other subclasses in Section~\ref{sec:res_cross}.

As shown in Fig.~\ref{fig:res_standard_paths}(a), the gain decreased with increasing $\delta_{\max}$.

For $\Delta c_{0.99}^{w}$, the gain did not change consistently in the negative direction (Fig.~\ref{fig:res_standard_paths}(b)).
In the positive direction, the gain remained similar up to S-$\Delta c$-pos-mid and decreased sharply at S-$\Delta c$-pos-high, for which negative transfer occurred (Fig.~\ref{fig:res_standard_paths}(b)).

For $\Delta H$, the gain did not change consistently for the targets deeper than the source (Fig.~\ref{fig:res_standard_paths}(c)).
For the targets shallower than the source, both the node and the element gains decreased as the forming depth decreased (Fig.~\ref{fig:res_standard_paths}(c)).
This may be attributed to the smaller deformation of shallower targets, which may make the target easier to learn from the target simulations alone.

Therefore, the gain changed in three regions: the full range of $\delta_{\max}$, the positive direction of $\Delta c_{0.99}^{w}$ and the negative direction of $\Delta H$.
For $\delta_{\max}$ and $\Delta H$, the gain decreased at each step of the variation.
In contrast, for $\Delta c_{0.99}^{w}$, the gain decreased only between S-$\Delta c$-pos-mid and S-$\Delta c$-pos-high.

The fourth descriptor was introduced based on the variation of $\Delta c_{0.99}^{w}$.
From S-$\Delta c$-zero to S-$\Delta c$-pos-mid, the gain did not change, while it decreased sharply from S-$\Delta c$-pos-mid to S-$\Delta c$-pos-high.
Since $\Delta c_{0.99}^{w}$ only describes the high-curvature regions of the die, the change in the overall amount of curvature was also examined.
For this purpose, the curvedness distribution distance $\delta_c$ was introduced, in mm$^{-1}$.
It is defined as the first Wasserstein distance between the curvedness distributions of the source and target die surfaces.
$\delta_c$ compares the two distributions without pairing the points by their positions.
Therefore, $\delta_c$ describes the change in the overall amount of curvature on the die, while $\Delta c_{0.99}^{w}$ describes the change in its high-curvature regions.
The values of $\Delta c_{0.99}^{w}$ and $\delta_c$ of S-$\Delta c$-zero, S-$\Delta c$-pos-mid and S-$\Delta c$-pos-high are listed in Table~\ref{tab:res_dc_values}.

\begin{table}[!htbp]
\centering
\caption{$\Delta c_{0.99}^{w}$, $\delta_c$ and K80 gain of the positive variation of $\Delta c_{0.99}^{w}$.}
\label{tab:res_dc_values}
\small
\begin{tabular}{@{}lccc@{}}
\toprule
Target & $\Delta c_{0.99}^{w}$ ($\times 10^{-3}$~mm$^{-1}$) & $\delta_c$ ($\times 10^{-4}$~mm$^{-1}$) & K80 gain (\%) \\
\midrule
S-$\Delta c$-zero & $-0.20$ & 0.87 & 26.2 \\
S-$\Delta c$-pos-mid & 2.00 & 2.68 & 25.9 \\
S-$\Delta c$-pos-high & 4.68 & 6.49 & $-15.3$ \\
\bottomrule
\end{tabular}
\end{table}

The increase in $\delta_c$ from S-$\Delta c$-pos-mid to S-$\Delta c$-pos-high was much larger than that from S-$\Delta c$-zero to S-$\Delta c$-pos-mid.
Therefore, the sharp decrease in gain at S-$\Delta c$-pos-high may be caused by the change in $\delta_c$ rather than by the change in $\Delta c_{0.99}^{w}$.

To study the effect of $\delta_c$, the other three descriptors should ideally be fixed or lie in a region where they have little effect on the gain.
Therefore, one additional geometry, S-$\delta_c$-high, was used together with two geometries from the previous variations, S-$\delta_c$-low and S-$\Delta c$-neg-high.
Their descriptors are listed in Table~\ref{tab:res_dc_path}, and their gains are shown in Fig.~\ref{fig:res_standard_paths}(d).
S-$\delta_c$-high and S-$\delta_c$-low have almost the same $\delta_{\max}$, $\Delta c_{0.99}^{w}$ and $\Delta H$, so the comparison between them shows the effect of $\delta_c$.
S-$\Delta c$-neg-high has an intermediate $\delta_c$.
Its $\Delta c_{0.99}^{w}$ differs from those of the other two geometries, but it lies in the negative direction of $\Delta c_{0.99}^{w}$, in which the gain did not change consistently (Fig.~\ref{fig:res_standard_paths}(b)).
Therefore, S-$\Delta c$-neg-high was used as an additional point to check the trend.

\begin{table}[!htbp]
\centering
\caption{Descriptors of the geometries in Fig.~\ref{fig:res_standard_paths}(d).}
\label{tab:res_dc_path}
\small
\begin{tabular}{@{}lcccc@{}}
\toprule
Target & $\delta_{\max}$ (mm) & $\Delta c_{0.99}^{w}$ ($\times 10^{-3}$~mm$^{-1}$) & $\Delta H$ (mm) & $\delta_c$ ($\times 10^{-4}$~mm$^{-1}$) \\
\midrule
S-$\delta_c$-low & 29.7 & $-0.20$ & 0.6 & 0.87 \\
S-$\Delta c$-neg-high & 29.8 & $-3.72$ & 0.6 & 1.95 \\
S-$\delta_c$-high & 29.8 & $-0.20$ & 0.6 & 3.37 \\
\bottomrule
\end{tabular}
\end{table}

Along the three geometries in Fig.~\ref{fig:res_standard_paths}(d), the gain decreased with increasing $\delta_c$, and S-$\delta_c$-high gave a much lower gain than S-$\delta_c$-low.
This indicates that $\delta_c$ also affects the transfer benefit of the geometries within the Standard subclass, and this effect is examined on the other subclasses in Section~\ref{sec:res_cross}.

\subsection{DISCOVER: comparison across subclasses}
\label{sec:res_cross}

\begin{figure}[!htbp]
\centering
\includegraphics[width=\textwidth]{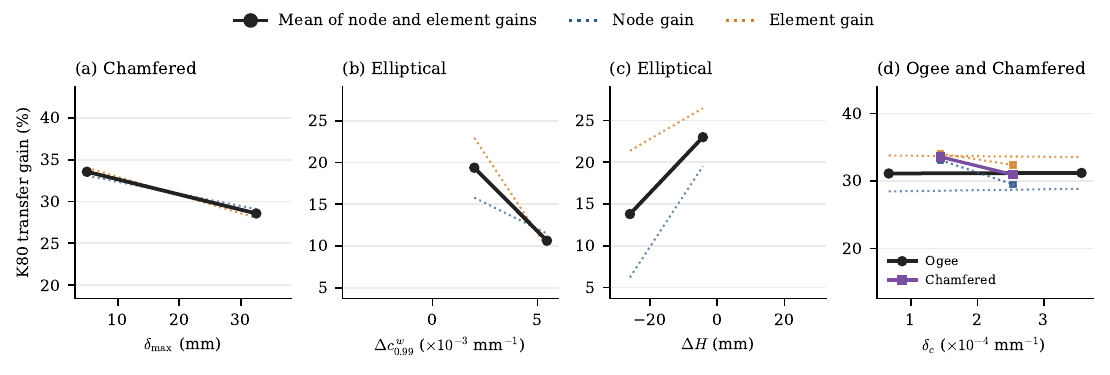}
\caption{K80 transfer gain of the Chamfered, Elliptical and Ogee geometries. (a) $\delta_{\max}$: C-$\delta_{\max}$-low and C-$\delta_{\max}$-high. (b) $\Delta c_{0.99}^{w}$: E-$\Delta c$-pos-mid and E-$\Delta c$-pos-high, which also differ in $\Delta H$. (c) $\Delta H$: E-$\Delta H$-neg-mid and E-$\Delta H$-neg-high. (d) $\delta_c$: O-$\delta_c$-low and O-$\delta_c$-high, and C-$\delta_c$-low and C-$\delta_c$-high. The solid lines show the mean of the node and element gains, and the dotted lines show the node and element gains. Each panel has its own vertical range, and all panels have the same span.}
\label{fig:res_cross}
\end{figure}

This subsection compares the trends observed in the Standard subclass with the results of the Chamfered, Elliptical and Ogee subclasses.
These geometries were not used to narrow the ranges of the descriptors in Section~\ref{sec:res_standard}, so they provide an independent check of the trends.
The gains of these geometries are shown in Fig.~\ref{fig:res_cross}, with the same horizontal axes as in Fig.~\ref{fig:res_standard_paths}.

For $\delta_{\max}$, $\Delta c_{0.99}^{w}$ and $\Delta H$, the gain of the other subclasses changed in the same direction as in the Standard subclass (Fig.~\ref{fig:res_cross}(a)--(c)).
The trend of $\Delta H$ agreed best, and the decrease in gain per unit change in $\Delta H$ was similar in the Standard and Elliptical subclasses.
For $\delta_{\max}$ and $\Delta c_{0.99}^{w}$, the decrease was smaller than in the Standard subclass.
For $\delta_c$, the gains of the Ogee geometries changed little, and so did the gains of C-$\delta_c$-low and C-$\delta_c$-high (Fig.~\ref{fig:res_cross}(d)).
Therefore, the decrease in gain with $\delta_c$ was observed only in the Standard subclass.

These results indicate that $\Delta H$, $\delta_{\max}$ and $\Delta c_{0.99}^{w}$ can be used to indicate target geometries for which a lower transfer benefit is expected, while the magnitude of the decrease depends on the subclass.
Therefore, for a new subclass, these relationships may be calibrated with a small number of target geometries before they are used to select target geometries.

\subsection{Overall transfer benefit}
\label{sec:res_overall}

\begin{figure}[!htbp]
\centering
\includegraphics[width=\textwidth]{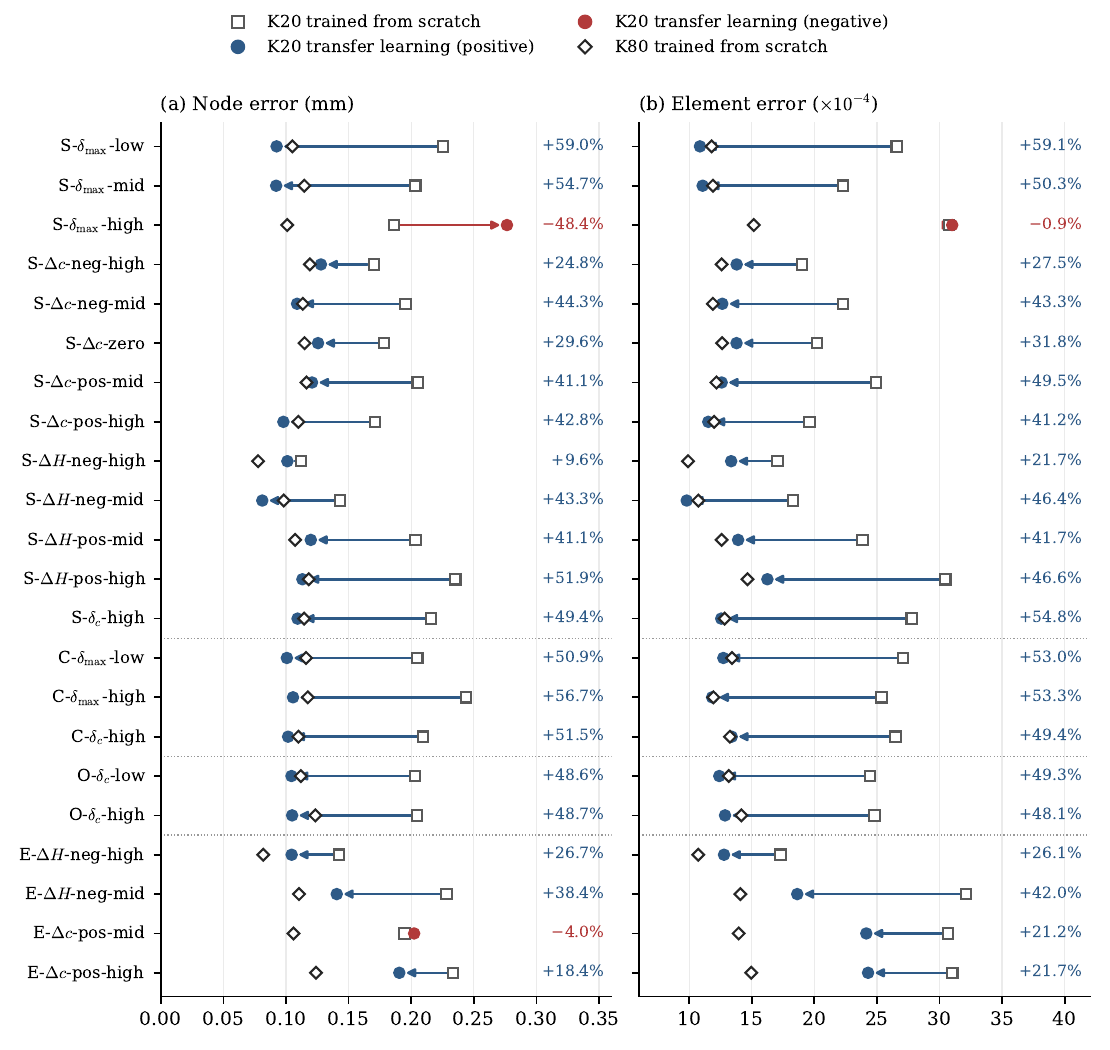}
\caption{Test errors of all target geometries for transfer learning with K20, training from scratch with K20 and training from scratch with K80. (a) Node error. (b) Element error. The arrows point from training from scratch to transfer learning with K20, and the values give the K20 transfer gain. The dotted lines separate the subclasses.}
\label{fig:res_efficiency}
\end{figure}

This subsection summarises the transfer benefit over the 22 target geometries of the systematic study.
Examples of the predicted fields after transfer learning are shown in Appendix~\ref{app:res_examples}.
At K80, the node and element gains were positive for 21 of the 22 target geometries, and negative transfer occurred only for S-$\Delta c$-pos-high.
Negative transfer at K80 also occurred for the node error of S-Max in Section~\ref{sec:res_transfer}, which has the largest Chamfer distance among the three selected target geometries.
At K20, transfer learning also gave a positive gain for most of the target geometries (Fig.~\ref{fig:res_efficiency}).
The largest negative transfer at K20 occurred for the node error of S-$\delta_{\max}$-high, which has the largest $\delta_{\max}$ among the target geometries, while its gain at K80 was positive.

\paragraph{Reduction in target data}
\label{sec:res_efficiency}
Fig.~\ref{fig:res_efficiency} also compares the transferred model with K20 and the model trained from scratch with K80.
A K20 error that was less than 5\% higher than the K80 error was regarded as similar.
With 20 target simulations, the transferred model gave errors similar to or lower than those of the model trained from scratch with 80 target simulations for about half of the target geometries, 13 of the 22 for the node error and 11 of the 22 for the element error.
This was the case for all Chamfered and Ogee geometries, but not for the Elliptical geometries.
For these geometries, transfer learning reduces the number of target simulations by 60 for each new global geometry.
In general, for $N_{\mathrm{g}}$ global geometries, the number of FE simulations required with the residual formulation and transfer learning is
\begin{equation}
N_{\mathrm{FE}} = N_{\mathrm{g}} + N_{\mathrm{src}} + \left(N_{\mathrm{g}} - 1\right) K,
\label{eq:res_doe_cost}
\end{equation}
where the first term denotes one baseline simulation for each global geometry, $N_{\mathrm{src}}$ denotes the number of rib simulations of the source geometry, and $K$ denotes the number of target simulations for each target geometry.
In contrast, a surrogate that predicts the complete fields for all combinations of $N_{\mathrm{g}}$ global geometries and $N_{\mathrm{l}}$ rib configurations may require a number of simulations that scales with $N_{\mathrm{g}} N_{\mathrm{l}}$.
Compared with transfer learning of a surrogate that predicts the complete fields, the residual formulation requires $N_{\mathrm{g}}$ additional baseline simulations.
Therefore, the number of FE simulations is mainly determined by $K$.
One FE simulation of the corner geometry takes approximately 1,600~s on an Intel Core i7-12700 CPU with 12 cores \cite{Zhao2026BiLoGo}.
Therefore, reducing $K$ from 80 to 20 saves approximately 26.7~h of FE computation for each target geometry, while the baseline simulation adds approximately 1,600~s.
Since the time of transfer learning is similar to the time of training from scratch, the saving in FE computation is the main reduction in computational cost.

\subsection{Limitations and future work}
\label{sec:res_limitations}

This subsection discusses the scope of the study and the directions for future work.
First, the transfer benefit was related only to the geometry, since the decision on transfer is usually needed before the target simulations are run.
Other factors, such as the difference between the source and target responses, were not quantified, and they may explain why the magnitude of the relationships differed between the subclasses.
In addition, the descriptors were defined explicitly so that they can be interpreted.
Descriptors learned from the surfaces or derived from the baseline response, combined with information from a small number of target simulations, may describe the transfer benefit more completely.
Second, the relationships are trends from controlled variations, and each setting was trained once.
Therefore, the conclusions are based on trends that are consistent across several target geometries, and they do not predict the magnitude of the gain for a new target geometry.
A model that predicts the gain from the descriptors may be developed with target geometries sampled across the subclasses.
Third, all transfers started from one source geometry and used one transfer strategy.
The choice of the source geometry, transfer strategies that depend on the change in global geometry, and a base model that grows with each new global geometry may reduce the negative transfer observed for large changes in global geometry.
Finally, the residual formulation requires one baseline FE simulation for each new global geometry, which may be replaced by a surrogate of the global geometry, such as BiLoGo \cite{Zhao2026BiLoGo}.
Other local features, such as beads and pads, and other process conditions may also be studied.

\section{Conclusions}
\label{sec:res_summary}

This study proposed a baseline-conditioned residual formulation and a surrogate, R-BiLoGo, to learn the effect of ribs on one global geometry and to transfer it to other global geometries of hot-stamped corner components.
The inputs of R-BiLoGo are the baseline response, which is the FE result of the same component without ribs, and the geometric perturbation, which describes the change in the tool surface caused by the ribs.
Its outputs are the changes in displacement and thinning from the baseline response.
The main findings are as follows.

First, R-BiLoGo learned the effect of the ribs accurately on the source geometry.
It gave a mean node error of 0.0743~mm, which was lower than those of a hierarchical graph-based model and a Transolver-based model.
The ablation study showed that the residual targets mainly reduced the node error, while the baseline response as input and the baseline fusion mainly reduced the thinning error.

Second, R-BiLoGo could be transferred to new global geometries with a small number of target simulations.
Without transfer learning, the node error of the source model on a new global geometry was 8.6 to 17.3 times that after transfer learning with 20 target simulations.
After transfer learning, R-BiLoGo gave lower errors than the two baseline models on two of three selected target geometries, while transfer learning of the Transolver-based model gave little or no benefit.
However, for the selected target geometry with the largest change in global geometry, transfer learning of R-BiLoGo gave higher node errors than training from scratch for up to 80 target simulations.

Third, transfer learning reduced the number of target simulations required.
Across the 22 target geometries of the systematic study, transfer learning with 20 target simulations gave errors similar to or lower than those of training from scratch with 80 target simulations for about half of the target geometries, including all Chamfered and Ogee geometries.
This saves approximately 26.7~h of FE computation for each target geometry.
With the same 80 target simulations, transfer learning reduced both the node and the element errors for 21 of the 22 target geometries.

Fourth, the benefit of transfer depended on the change in global geometry, which was described by four parameter-free geometric descriptors.
In the Standard subclass, the benefit decreased with increasing maximum surface distance $\delta_{\max}$, with increasing change in the high-curvature regions $\Delta c_{0.99}^{w}$ above a certain value, as the target became shallower than the source, and with increasing curvedness distribution distance $\delta_c$.
The trends of $\delta_{\max}$, $\Delta c_{0.99}^{w}$ and the forming-depth change $\Delta H$ were also observed in the Chamfered and Elliptical subclasses, and the trend of $\Delta H$ agreed best.
However, the magnitude of the decrease differed between the subclasses, and the decrease with $\delta_c$ was not observed in the Ogee and Chamfered subclasses.

These results show that the effect of local features learned on one global geometry can be reused on new global geometries with a small number of simulations.
In addition, the descriptors can be used to indicate target geometries for which a lower benefit of transfer is expected, and for a new subclass, these relationships may be calibrated with a small number of target geometries.
Future work may develop descriptors that describe the location and the form of the change in global geometry, and study the transfer of other types of local features.

\section*{Acknowledgement}
The authors acknowledge funding support from UKRI (UKRI221: AI-Driven Design for Forming High-Performance Vehicle Parts), as well as PhD scholarships from Imperial College London. The authors would like to thank Keysight for their technical support with the PAM-STAMP software. For the purpose of open access, the authors have applied a Creative Commons Attribution (CC BY) license to any Author Accepted Manuscript version arising.

\section*{CRediT authorship contribution statement}
\textbf{Yingxue Zhao}: Conceptualization, Methodology, Software, Investigation, Data curation, Formal analysis, Validation, Visualization, Writing -- original draft, Writing -- review \& editing.
\textbf{Haoran Li}: Methodology, Formal analysis, Validation, Writing -- review \& editing.
\textbf{Haosu Zhou}: Methodology, Formal analysis, Writing -- review \& editing.
\textbf{Tobias Pfaff}: Methodology, Validation, Supervision, Writing -- review \& editing.
\textbf{Nan Li}: Conceptualization, Methodology, Formal analysis, Validation, Supervision, Project administration, Funding acquisition, Writing -- review \& editing.

\section*{Declaration of competing interest}
The authors declare that they have no known competing financial interests or personal relationships that could have appeared to influence the work reported in this paper.

\section*{Data availability}
The data supporting the findings of this study are available upon reasonable request.

\appendix

\section{Selection and control of geometric descriptors}
\label{app:res_descriptors}

\subsection{Descriptor selection}
\label{subsec:descriptor_selection}
\label{app:descriptor_measurement}

The candidate descriptors cover five types of geometric difference between the source and target forming surfaces: surface position, surface orientation, curvature, cross-sectional profile and global dimensions.
Each candidate combines a geometric field, such as the distance between the two surfaces, with a statistic, such as the mean, a percentile or the maximum.
In total, 28 candidates were defined.
All candidates were evaluated in the common manufacturing coordinate system of the CAD models, without a best-fit rigid registration, since such a registration could remove the changes in position and forming depth that should be measured.
A technical pre-screening of the definition, the computability and the convergence with the surface discretisation left 15 candidates.

The 15 candidates were computed on 430 CAD geometries: 130 Standard geometries and 100 geometries for each of the other three subclasses.
All descriptors were computed relative to the same Standard source geometry.
The geometries included both controlled variations of single CAD parameters and space-filling samples.
For descriptor $j$ and subclass $f$, the response within the subclass was measured by
\begin{equation}
S_{j,f}=\frac{Q_{0.95}(D_j\mid f)-Q_{0.05}(D_j\mid f)}{R_{j,\mathrm{pool}}},
\label{eq:app_descriptor_response}
\end{equation}
where $D_j$ denotes the value of descriptor $j$, $Q_p$ denotes the $p$-quantile, and $R_{j,\mathrm{pool}}$ denotes the same range over all 430 geometries.
A descriptor was required to respond within each subclass, rather than mainly between the subclasses.
The redundancy between descriptors $j$ and $k$ was measured by
\begin{equation}
d_{jk}^{\mathrm{red}}=\max\left\{\operatorname{dCor}(D_j,D_k),\operatorname{dCor}(|D_j|,|D_k|)\right\},
\label{eq:app_redundancy_measure}
\end{equation}
where $\operatorname{dCor}$ denotes the distance correlation.
The magnitude-based term is included so that two signed descriptors with strongly related magnitudes are not treated as independent.

Based on these two criteria, three descriptors were retained,
\begin{equation}
\boldsymbol{\gamma}(S,T)=
\begin{bmatrix}
\delta_{\max}(S,T) &
\Delta c_{0.99}^{w}(S,T) &
\Delta H(S,T)
\end{bmatrix}^{\top}.
\label{eq:geometry_descriptor_vector}
\end{equation}
They describe the local positional difference, the change in the high-curvature regions and the change in forming depth, respectively.
Their pairwise redundancy measures were at most 0.41.
In contrast, a mean surface distance responded strongly to the geometry, but it was almost fully coupled to $\Delta H$, with $d^{\mathrm{red}}=0.997$, and it was therefore not retained.

\subsection{Descriptor definitions}

Let $\mathcal{X}_S$ and $\mathcal{X}_T$ denote the points sampled on the source and target forming surfaces, with 32,768 points on each surface.
For a point $\mathbf{x}\in\mathcal{X}_S$, let $\pi_T(\mathbf{x})$ denote its reconstructed location on the target CAD surface, and let $\pi_S$ be defined in the same way.
The sampled maximum bidirectional surface distance is
\begin{equation}
\delta_{\max}(S,T)=\max\left\{\max_{\mathbf{x}\in\mathcal{X}_S}\left\|\mathbf{x}-\pi_T(\mathbf{x})\right\|_2,\;\max_{\mathbf{y}\in\mathcal{X}_T}\left\|\mathbf{y}-\pi_S(\mathbf{y})\right\|_2\right\}.
\label{eq:sampled_maximum_distance}
\end{equation}

The curvedness at a point of the surface is
\begin{equation}
c=\sqrt{\frac{\kappa_1^2+\kappa_2^2}{2}},
\label{eq:curvedness_definition}
\end{equation}
where $\kappa_1$ and $\kappa_2$ denote the principal curvatures, computed from the CAD surface \cite{KoenderinkVanDoorn1992}.
The area-weighted 99th-percentile curvedness change is
\begin{equation}
\Delta c_{0.99}^{w}(S,T)=Q_{0.99}^{w}(c_T)-Q_{0.99}^{w}(c_S),
\label{eq:curvedness_quantile_change}
\end{equation}
where $Q_{0.99}^{w}$ denotes the area-weighted 99th percentile, and $c_S$ and $c_T$ denote the curvedness fields of the source and target surfaces.
A positive value indicates more high-curvature regions on the target surface.

The forming-depth change is
\begin{equation}
\Delta H(S,T)=H(T)-H(S),
\label{eq:forming_depth_change}
\end{equation}
where $H$ denotes the forming depth.
A negative value indicates that the target is shallower than the source.

\subsection{Generation of controlled target geometries}
\label{subsec:descriptor_control}
\label{app:descriptor_control}

A change in one CAD parameter usually changes several descriptors.
Therefore, the controlled target geometries were generated with a radial basis function (RBF) surrogate of the mapping from the CAD parameters to the descriptors \cite{WildRegisShoemaker2008}.
For the Standard subclass, the CAD parameters are the die radius, the punch radius, the plan-view radius and the forming depth, which were min--max normalised to $\mathbf{x}\in[0,1]^4$.
The surrogate
\begin{equation}
\widehat{\boldsymbol{\gamma}}(\mathbf{x})\approx\boldsymbol{\gamma}\left(S,T(\mathbf{x})\right)
\label{eq:rbf_descriptor_mapping}
\end{equation}
was fitted to 128 Standard geometries, where $T(\mathbf{x})$ denotes the target geometry generated with the parameters $\mathbf{x}$.
For a prescribed descriptor vector $\mathbf{t}$, the CAD parameters were found by
\begin{equation}
\mathbf{x}^{*}=\arg\min_{\mathbf{x}\in[0,1]^4}\sum_d\left(\frac{\widehat{\gamma}_d(\mathbf{x})-t_d}{s_d}\right)^2,
\label{eq:descriptor_target_matching}
\end{equation}
where $s_d=Q_{0.95,d}-Q_{0.05,d}$ denotes the range of descriptor $d$ in the Standard geometries.
For each descriptor, the low and high cases used its 10th and 90th percentiles, while the other two descriptors were set to their medians.
Subsequent searches for the Chamfered, Elliptical and Ogee subclasses fitted separate RBF surrogates to measured geometries of the respective subclass.
The proposed geometries were measured before the final selection.

For the high-$\delta_{\max}$ case, the RBF result did not give a sufficient increase in $\delta_{\max}$.
This case was refined by a null-space search \cite{Rosen1960}.
The gradient of $\delta_{\max}$ was projected onto the null space of the Jacobian $\widehat{J}_c$ of the other two descriptors,
\begin{equation}
\mathbf{v}_{\delta}=\left(I-\widehat{J}_c^{\dagger}\widehat{J}_c\right)\nabla_{\mathbf{x}}\widehat{\gamma}_{\delta},
\label{eq:nullspace_projection}
\end{equation}
where $(\cdot)^{\dagger}$ denotes the pseudoinverse.
Moving along $\mathbf{v}_{\delta}$ increases $\delta_{\max}$ while the first-order changes in $\Delta c_{0.99}^{w}$ and $\Delta H$ are zero.
The Jacobian was recomputed along the search, since the mapping is nonlinear.
With this refinement, $\delta_{\max}$ increased from 32.9~mm to 36.5~mm, while $\Delta H$ and $\Delta c_{0.99}^{w}$ remained almost unchanged.

\begin{figure}[!htbp]
\centering
\includegraphics[width=\textwidth]{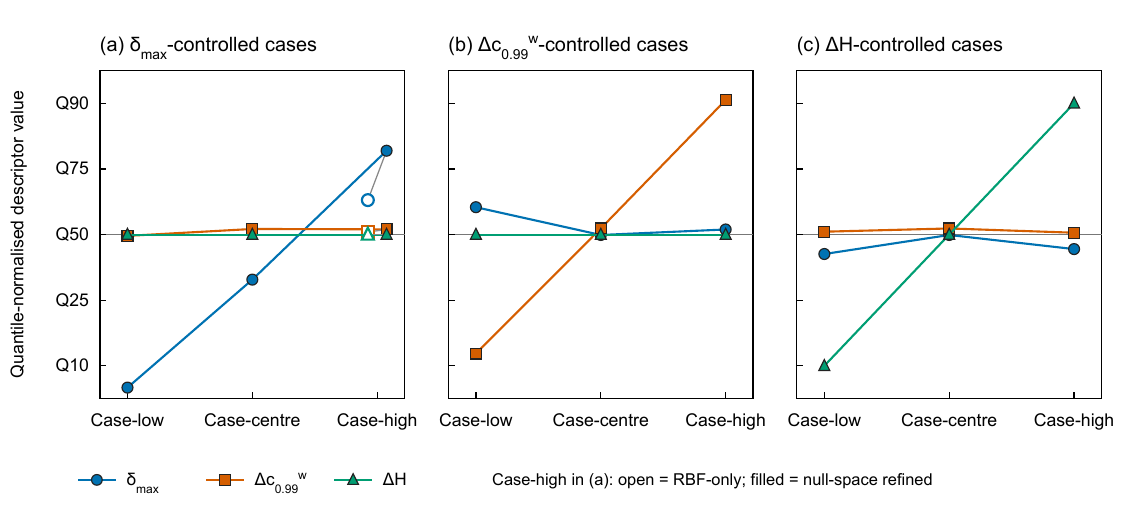}
\caption{Controlled variations of (a) $\delta_{\max}$, (b) $\Delta c_{0.99}^{w}$ and (c) $\Delta H$, measured from the regenerated CAD geometries. Circles, squares and triangles denote $\delta_{\max}$, $\Delta c_{0.99}^{w}$ and $\Delta H$. The vertical axis shows the quantile of each descriptor in the 128 Standard geometries. In (a), the open markers at the high case show the result before the null-space refinement.}
\label{fig:app_descriptor_control}
\end{figure}

All selected parameter sets were regenerated as CAD geometries, and the descriptors were measured again from the realised surfaces.
Therefore, all descriptor values used in this study are measured values, not RBF predictions.
The resulting controlled variations are shown in Fig.~\ref{fig:app_descriptor_control}.

\section{Nested subsets of rib configurations}
\label{app:res_subsets}

\begin{figure}[!htbp]
\centering
\includegraphics[width=\textwidth]{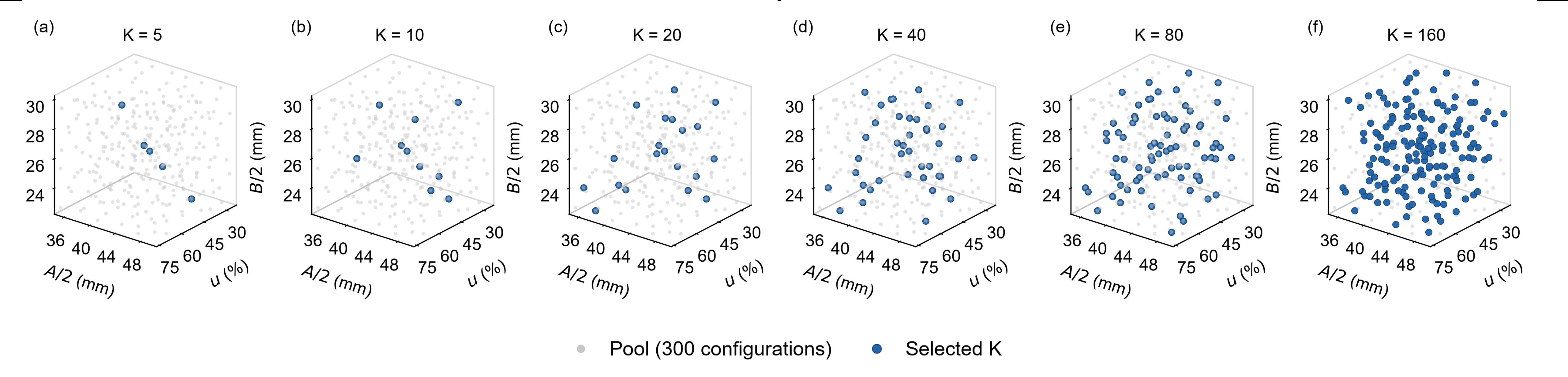}
\caption{Nested subsets of rib configurations for (a) $K = 5$, (b) $K = 10$, (c) $K = 20$, (d) $K = 40$, (e) $K = 80$ and (f) $K = 160$. The grey points show the pool of 294 rib configurations, and the blue points show the selected configurations.}
\label{fig:res_subsets}
\end{figure}

The nested subsets of Eq.~\eqref{eq:res_subsets} were selected with a deterministic method based on a Halton sequence, which is a low-discrepancy sequence that covers the unit cube evenly \cite{Halton1960Sequence}.
First, the three rib parameters $u$, $A/2$ and $B/2$ of the pool were normalised to $[0,1]$.
A three-dimensional Halton sequence with a fixed random shift \cite{Cranley1976Randomization} was then generated.
At each step, the unused configuration of the pool that is closest to the next point of the sequence was selected.
A small penalty was added to this distance, which is proportional to the number of previously selected configurations that fall into the same one-dimensional interval of each parameter.
This penalty prevents the selected configurations from concentrating in the same range of one parameter.
This gives a fixed ranking of the pool, and $S_K$ contains the first $K$ configurations of this ranking.
Only the rib parameters were used for the selection, and no simulation results were used.
The selected subsets are shown in Fig.~\ref{fig:res_subsets}.
Each subset covers the rib parameter space, and each larger subset contains the smaller ones.

\section{Learning curves of R-Transolver}
\label{app:res_transolver}

\begin{figure}[!htbp]
\centering
\includegraphics[width=\textwidth]{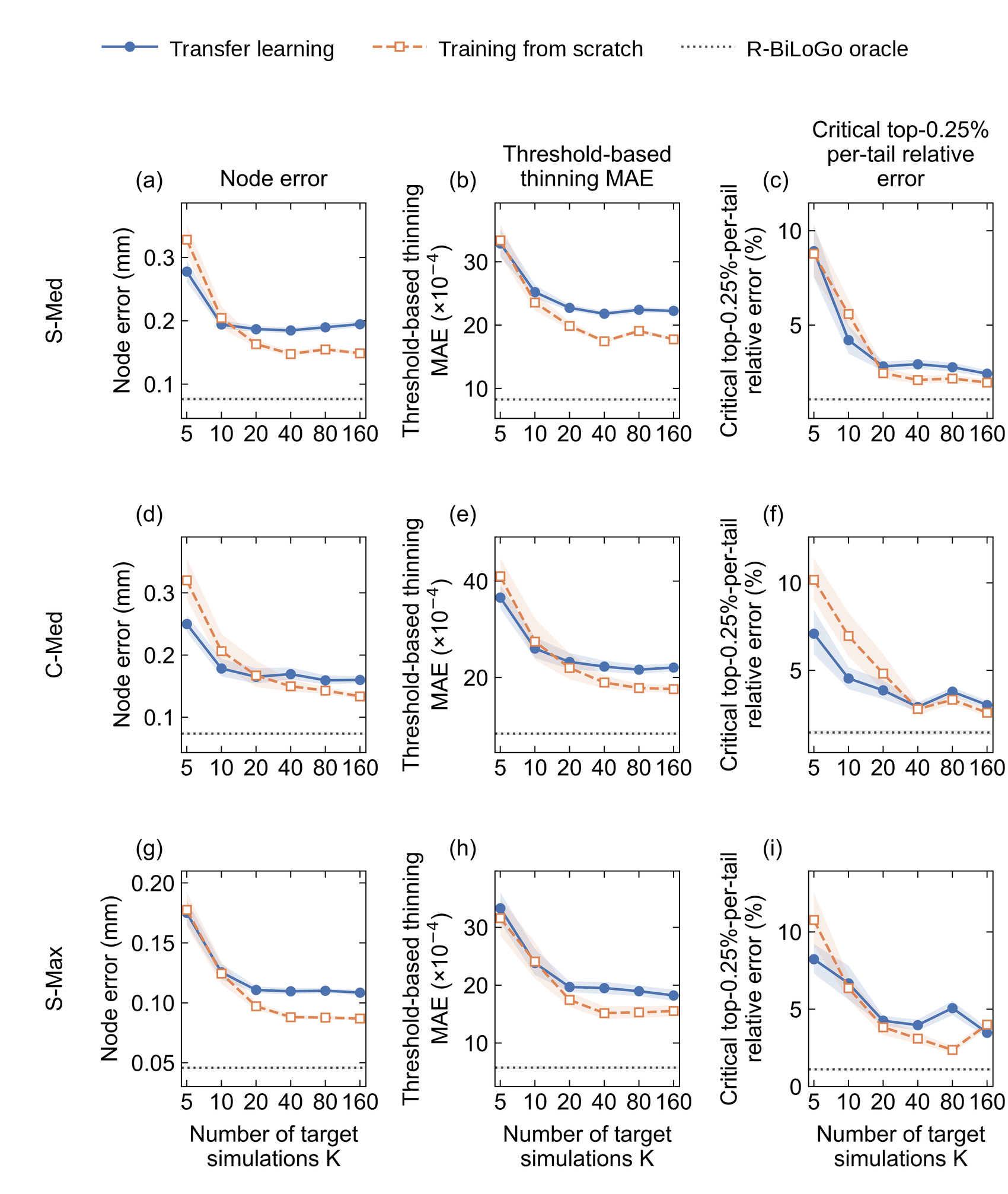}
\caption{Test errors of R-Transolver against the number of target simulations $K$ for transfer learning (solid lines) and training from scratch (dashed lines). The shaded bands show the standard deviation of the errors over the 50 test simulations. The dotted lines show the errors of the R-BiLoGo oracle, which was trained from scratch on the full training pool of 294 simulations. Rows: S-Med, C-Med and S-Max. Columns: mean positional Euclidean error, threshold-based thinning error and top-0.25\% critical relative thinning/thickening error.}
\label{fig:res_transolver_learning_curves}
\end{figure}

Fig.~\ref{fig:res_transolver_learning_curves} shows the test errors of R-Transolver against the number of target simulations $K$ on S-Med, C-Med and S-Max, for transfer learning from the source model and training from scratch.
The dotted lines show the errors of the R-BiLoGo oracle in Fig.~\ref{fig:res_learning_curves}.
After transfer learning, the errors of R-Transolver decreased from $K = 5$ to $K = 20$, and hardly decreased for $K \ge 20$.
For $K \ge 40$, training R-Transolver from scratch gave lower node and thinning errors than transfer learning for all three targets.
For all values of $K$, the errors of R-Transolver remained higher than those of the R-BiLoGo oracle.

\section{Examples of predictions after transfer learning}
\label{app:res_examples}

\begin{figure}[!htbp]
\centering
\includegraphics[width=\textwidth]{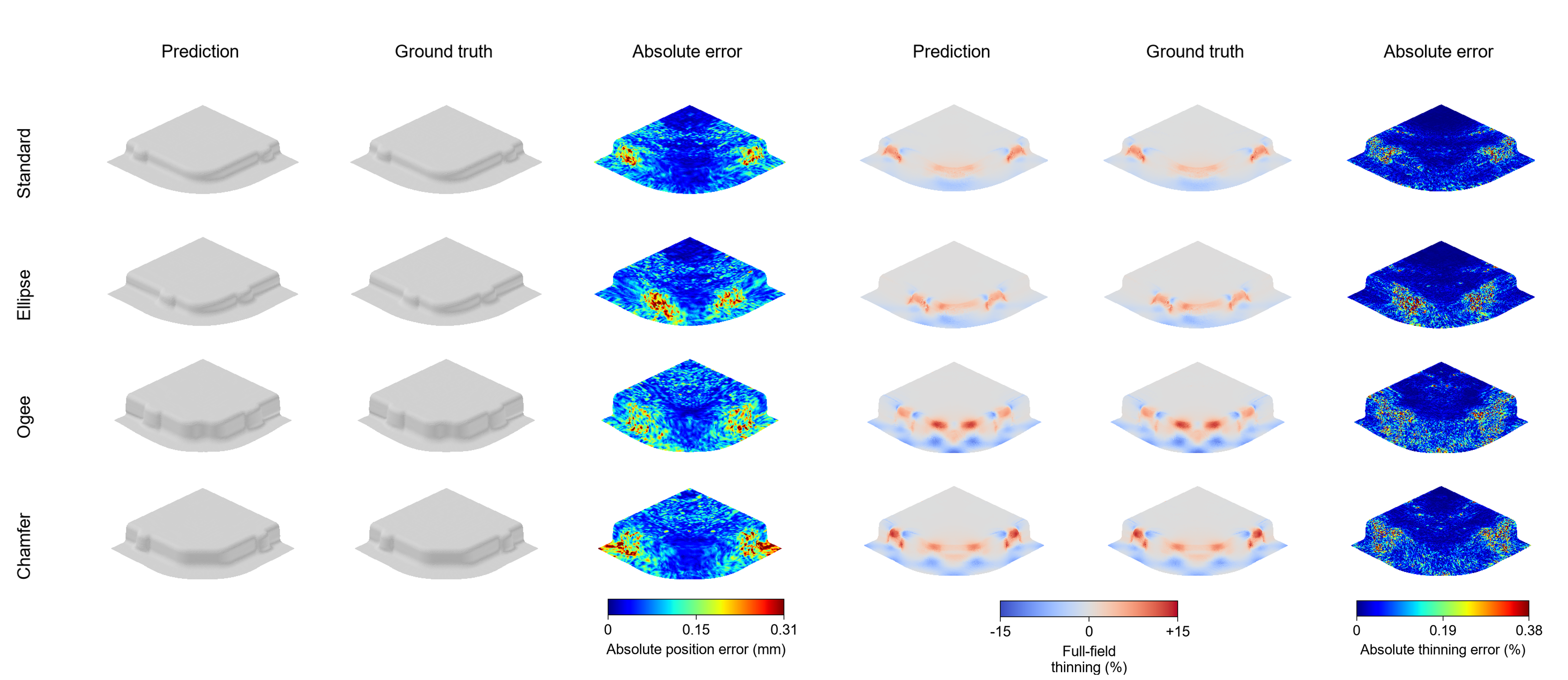}
\caption{Predictions of R-BiLoGo after transfer learning with K80 for one target geometry of each subclass. From left to right: predicted final geometry, FE final geometry and absolute position error, and predicted thinning, FE thinning and absolute thinning error.}
\label{fig:res_transfer_examples}
\end{figure}

Fig.~\ref{fig:res_transfer_examples} shows the predictions after transfer learning with K80 for one target geometry of each subclass, with the ribs of one test simulation.
The predicted final geometries and thinning fields agree closely with the FE results, and the larger errors occur mainly near the ribs and on the side walls.

\bibliographystyle{unsrtnat}
\bibliography{references}
\end{document}